\documentclass[10pt,twocolumn,letterpaper]{article}

\usepackage[pagenumbers]{cvpr} %

\usepackage{pifont}
\usepackage{xcolor}
\usepackage{algorithmicx} 
\usepackage{algorithm}
\usepackage{algpseudocode} 

\newcommand{\cmark}{\textcolor{green!50!black}{\checkmark}}   %
\newcommand{\xmark}{\textcolor{red!70!black}{\ding{55}}}      %
\usepackage{xspace}
\newcommand{\METHOD}{{QuPAINT\xspace}}

\definecolor{cvprblue}{rgb}{0.21,0.49,0.74}
\usepackage[pagebackref,breaklinks,colorlinks,allcolors=cvprblue]{hyperref}

\def\paperID{*****} %
\def\confName{CVPR}
\def\confYear{2026}

\title{QuPAINT: Physics-Aware Instruction Tuning Approach to \\ Quantum Material Discovery
}

\author{Xuan-Bac Nguyen$^{1}$, Hoang-Quan Nguyen$^{1}$, Sankalp Pandey$^{1}$, Tim Faltermeier$^{2}$, \\ Nicholas Borys$^{2}$, Hugh Churchill$^{3}$, Khoa Luu$^{1}$\\
    $^{1}$ CVIU Lab, University of Arkansas, USA \quad 
	$^{2}$ University of Utah, USA \\
        $^{3}$ Department of Physics, University of Arkansas, USA \\
	\tt\small $^{1}$\{xnguyen, hn016, sankalpp, khoaluu\}@uark.edu,\\ \tt\small$^{2}$\{timfaltermeier@montana.edu,nicholas.borys@utah.edu\}, \tt\small$^{3}$hchurch@uark.edu \\
    \small\url{https://uark-cviu.github.io/projects/qupaint}
}

\begin{document}

\twocolumn[{
\renewcommand\twocolumn[1][]{#1}%
\maketitle
\begin{center}
    \vspace{-8mm}
    \captionsetup{type=figure}
    \includegraphics[width=0.95\textwidth]{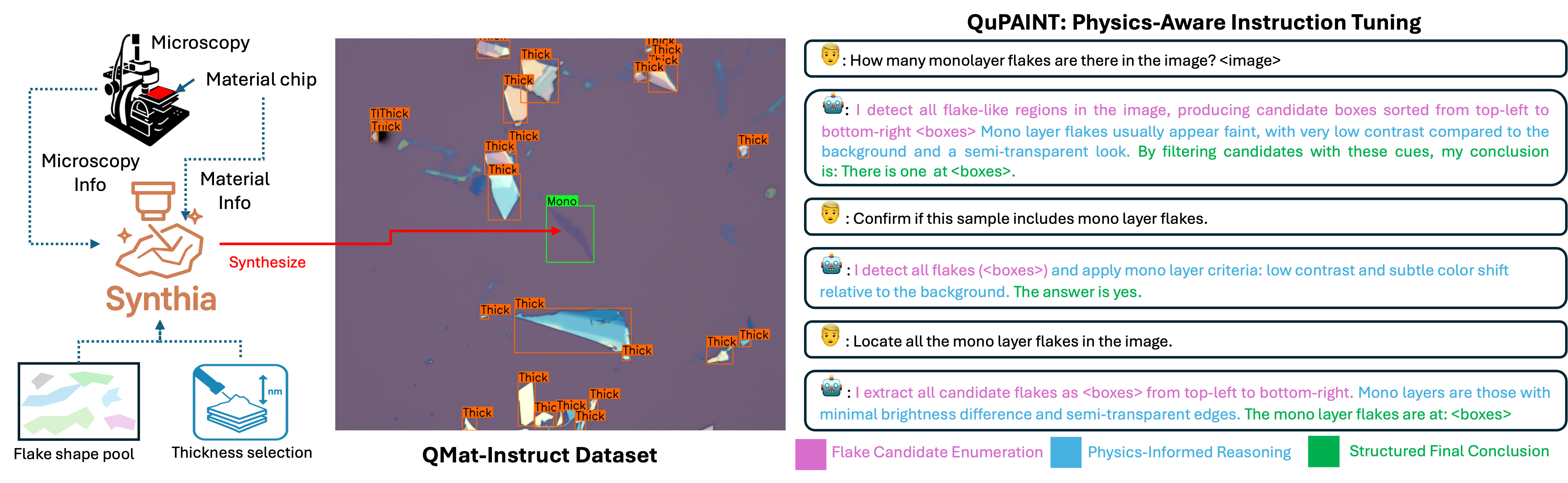}
    \vspace{-3mm}
    \captionof{figure}{
    Overview of our proposed system. \textbf{Synthia} generates physics-based synthetic flakes, and \textbf{QMat-Instruct} provides multimodal, physics-aware supervision. Together, they enable the new Physics-Aware Instruction Tuning for Quantum Material Discovery (\textbf{\METHOD}) framework, to learn robust and interpretable representations for quantum material characterization. \textbf{(Best view in colors)}
    }
\label{fig:teaser}
\end{center}
}]

\begin{abstract}
Characterizing two-dimensional quantum materials from optical microscopy images is challenging due to the subtle layer-dependent contrast, limited labeled data, and significant variation across laboratories and imaging setups. Existing vision models struggle in this domain since they lack physical priors and cannot generalize to new materials or hardware conditions. This work presents a new physics-aware multimodal framework that addresses these limitations from both the data and model perspectives. We first present Synthia, a physics-based synthetic data generator that simulates realistic optical responses of quantum material flakes under thin-film interference. Synthia produces diverse and high-quality samples, helping reduce the dependence on expert manual annotation. We introduce QMat-Instruct, the first large-scale instruction dataset for quantum materials, comprising multimodal, physics-informed question–answer pairs designed to teach Multimodal Large Language Models (MLLMs) to understand the appearance and thickness of flakes. Then, we propose Physics-Aware Instruction Tuning (\METHOD), a multimodal architecture that incorporates a Physics-Informed Attention module to fuse visual embeddings with optical priors, enabling more robust and discriminative flake representations. Finally, we establish QF-Bench, a comprehensive benchmark spanning multiple materials, substrates, and imaging settings, offering standardized protocols for fair and reproducible evaluation. 
\end{abstract}
    
\vspace{-7mm}
\section{Introduction}
\label{sec:intro}

\begin{figure*}[!ht]
    \centering
    \includegraphics[width=0.9\linewidth]{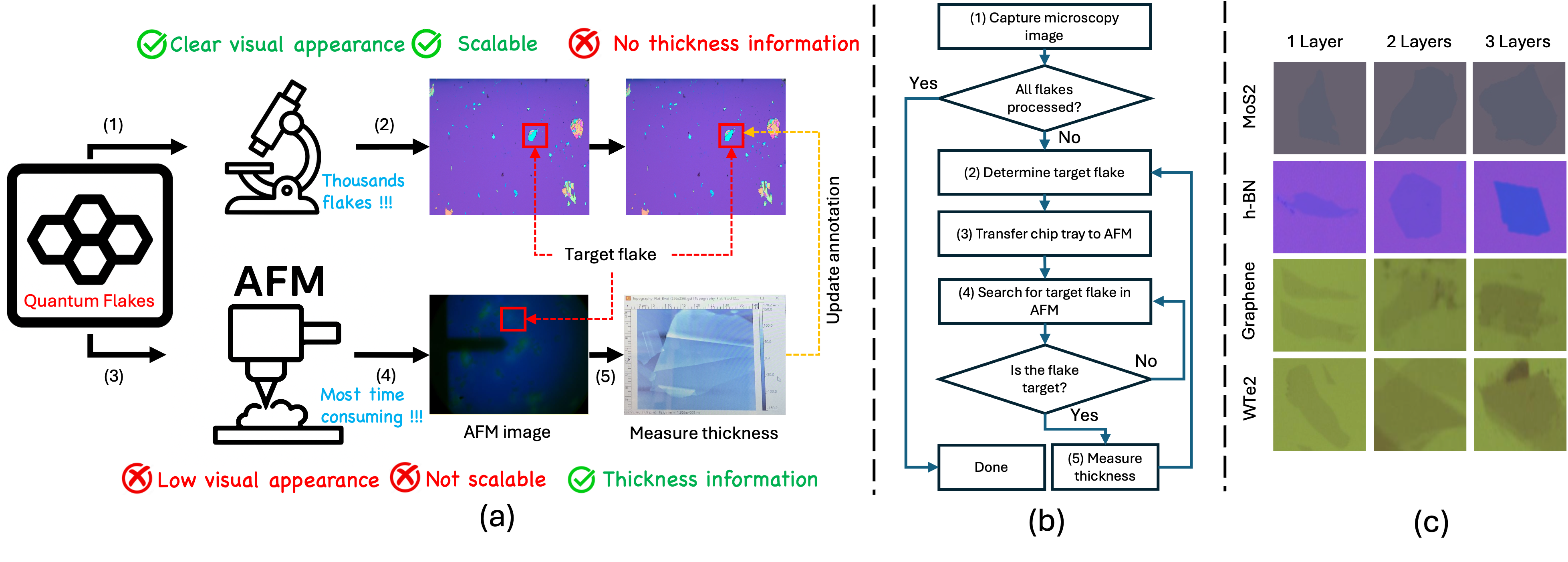}
    \vspace{-6mm}
    \caption{
    \textbf{Overview of the real data collection pipeline and challenges in quantum flake characterization.}
    (a) Microscopy enables fast collection of thousands of flakes but provides no thickness information, while AFM offers accurate layer measurements but is slow and not scalable. 
    (b) The required manual workflow is labor-intensive, involving repeated chip transfers and flake verification. 
    (c) Real samples from multiple materials (MoS$_2$, h-BN, Graphene, WTe$_2$) show minimal visual differences between 1, 2, ` 3 layers, illustrating the difficulty of determining thickness from raw optical images. \textbf{(Best view in colors)}}
    \label{fig:dataset_challenging}
    \vspace{-4mm}
\end{figure*}

Quantum materials, particularly atomically thin two-dimensional (2D) systems such as graphene, $\text{MoS}_2$, and $\text{WTe}_2$, have emerged as a new frontier for next-generation electronics, photonics, and quantum information technologies. Their extraordinary physical properties, such as quantum confinement, spin–valley coupling, and layer-dependent band topology, arise from atomic-scale variations that are invisible to conventional imaging and analysis pipelines \cite{ISLAM2025100161}. Characterizing these materials precisely, especially determining the thickness, uniformity, and local defect structures of exfoliated flakes, is therefore a critical step toward harnessing their quantum functionalities. Yet, this characterization process remains largely manual, subjective, and inconsistent across laboratories, limiting reproducibility and scalability. From a computer vision perspective, quantum material characterization presents a uniquely challenging regime: the visual differences between mono-, bi-, and few-layer flakes can be subtler than illumination noise, and their optical contrast often entangles with background interference patterns.

\noindent
\noindent\textbf{Limitations of Object Detection Methods. }
Despite the impressive progress of foundational CNN architectures \cite{he2017mask, redmon2016you} to recent transformer-based systems \cite{robinson2025rfdetrneuralarchitecturesearch, ravi2024sam, peng2024dfineredefineregressiontask, Zhou_2025_ICCV, assran2025vjepa2selfsupervisedvideo, siméoni2025dinov3}, these models remain insufficient for quantum material characterization. Conventional object detectors are trained to localize and classify semantically distinct objects, where category boundaries are defined by strong texture, color, or shape cues. In contrast, 2D quantum materials exhibit sub-visual variations: two flakes may appear nearly identical to a detector but correspond to entirely different physical phases, e.g., monolayer vs. bilayer $\text{MoS}_2$ (\cref{fig:dataset_challenging} (c)). Their optical appearances are easily distorted by microscope illumination, substrate thickness, or ambient humidity, breaking the visual assumptions of existing models. Moreover, unlike natural images, where large annotated datasets enable domain generalization, quantum flake datasets are extremely scarce and exhibit high intra-class variability that makes end-to-end training of state-of-the-art architectures prone to overfitting. Finally, the task itself goes beyond mere detection: researchers require physically interpretable predictions, such as layer count or thickness gradient, quantities that standard vision models are not designed to infer. These challenges call for a new framework that couples physical priors with visual representations, enabling models to reason jointly about material appearance and its underlying quantum properties.

\noindent\textbf{Limitations in Prior Work. } While several recent works have demonstrated promising deep-learning pipelines for identifying exfoliated flakes under optical microscopes \cite{masubuchi2020deep, uslu2024open, uslu2025maskterial}, their practical impact remains limited. Each study typically collects its own in-house dataset, focusing on specific substrates, e.g., $\text{SiO}_2/\text{Si}$ thickness, lighting conditions, or material types, and trains models that perform well only within that narrow domain. However, in real laboratory settings, imaging configurations vary substantially across institutions: optical magnification, illumination spectra, camera sensors, and even environmental factors can alter color contrast and scattering patterns. As a result, models overfit to their acquisition pipelines and fail to generalize to unseen microscopes or new materials of interest. Moreover, the community still lacks a standardized benchmark for quantum material vision, making it impossible to fairly compare methods or quantify their transferability in practice. Without unified datasets or evaluation protocols, progress in this area remains fragmented and difficult to reproduce. This gap motivates the need for a physically grounded, domain-adaptive framework that can learn transferable representations across labs, materials, and imaging modalities. toward reliable, real-world deployment of autonomous quantum material characterization systems.

\noindent\textbf{Contributions of this Work.} In summary, this work has four main contributions.
First, we introduce Synthia, a synthetic data generation framework that simulates realistic quantum material flakes using physics-based optical modeling. Synthia produces diverse, high-quality samples that help mitigate the limited availability of expert-labeled microscopy data.
Second, we construct QMat-Instruct, the first large-scale dataset of instructions for quantum materials. It provides multimodal, physics-aware question–answer pairs that teach an MLLM to reason about flake appearance and thickness.
Third, we propose the Physics-Aware Instruction Tuning (\METHOD), a new multimodal framework that integrates a Physics-Informed Attention module. It leverages optical priors and interference-aware cues to enable the model to learn more reliable and discriminative flake representations. \cref{fig:teaser} illustrates an overview of our proposed method.
Finally, we establish QF-Bench, a large-scale benchmark for quantum material characterization, covering multiple materials, substrates, and imaging settings. This benchmark provides unified evaluation protocols and allows fair comparison across future methods.

\section{Related Work}

\noindent\textbf{Automated Two-Dimensional Material Characterization.}
Research on 2D materials has grown rapidly due to their importance in quantum technologies \cite{dendukuri2020definingquantumneuralnetworks,nguyen2023quantum, Ouaj_2024, Ouaj_2023, steiner2025current, mckenzie2024fabrication, jackering2025super, PALINKAS2024119608, Hattori_2024, wegerhoff2025coherent, crimmann2025high, uslu2025maskterial, courtney2025automated}. Early deep-learning approaches include UNet-based segmentation \cite{ronneberger2015u, han2020deep} and Mask R-CNN pipelines for flake detection \cite{he2017mask, masubuchi2020deep}. Other work explores self-attention with soft-labeling \cite{nguyen2024two} and false-negative mitigation \cite{luu2024automatically}. Traditional machine-learning models such as Gaussian mixture models \cite{uslu2024open} have also been used but are sensitive to noise and perform poorly on real data. High-throughput microscopy systems aim to scale identification and enable statistical analysis \cite{crimmann2025high}. Foundation models trained on 2D material data \cite{uslu2025maskterial} and segmentation models like SAM \cite{kirillov2023segment, ravi2024sam} provide strong priors for downstream tasks. Recent studies integrate detection with fabrication workflows for improving exfoliation, sample handling, and labeling consistency \cite{mckenzie2024fabrication, Ouaj_2024, Ouaj_2023, jackering2025super, courtney2025automated}.

\noindent\textbf{Multimodal Large Language Models (MLLMs).}
MLLMs extend LLM reasoning to images and audio. Models such as CLIP \cite{radford2021learning} and ALIGN \cite{jia2021scaling} learn joint embeddings, while instruction-tuned systems like BLIP-2 \cite{li2023blip}, LLaVA \cite{liu2023visual}, and InstructBLIP \cite{dai2023instructblip} connect visual encoders to pretrained LLMs for open-ended reasoning. Follow-up work improves step-by-step reasoning \cite{xu2025llava}, multi-agent collaboration \cite{dong2025insight}, and task-specific alignment \cite{yan2025task, tan2025beyond}. Large-scale systems, e.g., GPT-4V \cite{achiam2023gpt}, Qwen-VL \cite{bai2023qwen}, and InternVL \cite{chen2024internvl} further expand capabilities to videos. Temporal modeling is another emerging direction. State Space Models aid long-sequence motion generation \cite{zhang2024motion}, while video understanding has improved through temporal reasoning of events \cite{qian2024momentor}. Frame selection strategies also boost long-form video performance without retraining \cite{liu2025bolt}. 
Despite these advances, vision-centric models still \textit{struggle to generalize across domains}. Our work addresses this limitation by introducing a physics-informed multimodal framework that links visual cues with quantum material properties for more reliable characterization.

\section{Challenges and Limitations of Prior Dataset}
\subsection{Challenges In Data Collection} 
The exfoliation of two-dimensional (2D) quantum materials, such as graphene and transition-metal dichalcogenides (TMDs), typically relies on the mechanical cleavage method using adhesive tape which is an approach recognized by the 2010 Nobel Prize in Physics. In this process, bulk crystals are repeatedly peeled to obtain atomically thin flakes, which are then transferred onto SiO$_2$/Si substrates. Although this technique is simple and inexpensive, it is inherently stochastic; the resulting flakes exhibit highly variable shapes, sizes, and thicknesses. 

\begin{table}[!t]
\centering
\caption{Comparison of dataset synthesis techniques between MaskTerial~\cite{uslu2025maskterial} and our proposed method.}
\label{tab:synthesis_comparison}
\vspace{-3mm}
\resizebox{\columnwidth}{!}{
\begin{tabular}{lcc}
\toprule
\textbf{Technique Aspect} & \textbf{MaskTerial~\cite{uslu2025maskterial}} & \textbf{Synthia (Ours)} \\
\midrule
Flake shapes from multi-material sources & \cmark & \cmark \\
Physics-based optical simulation (TMM) & \cmark & \cmark \\
Personalized color calibration & \xmark & \cmark \\
Using CIE standard & \xmark & \cmark \\
Avoid overlapping with the existing flakes & \xmark & \cmark \\
\bottomrule
\end{tabular}}
\vspace{-6mm}
\end{table}

Collecting large-scale data of exfoliated flakes is, in principle, straightforward since thousands of random flakes can be produced in a single exfoliation attempt. However, generating datasets that contain flakes with specific layer numbers or thicknesses, e.g., monolayer, bilayer, or few-layer regions, is extremely challenging. The randomness of the exfoliation process prevents direct control over the number of layers, meaning that flakes of interest must be ``hunted'' from among a vast background of irrelevant samples. 

This hunting process is extraordinarily time-consuming. The optical microscope images alone cannot determine layer thickness precisely because color contrast varies nonlinearly with illumination and substrate conditions. Accurate thickness measurement requires atomic force microscopy (AFM), which introduces substantial labor overhead: each candidate flake must be physically relocated to the AFM, measured at the nanometer scale, and then carefully remapped to its original microscopy coordinates. Such cross-modality alignment---between optical and AFM images---demands extensive manual effort and often limits the throughput of data collection. 
As a result, constructing quantum flake datasets suitable for AI training is both labor-intensive and data-scarce. The combination of random flake generation, uncertain thickness labeling, and slow AFM-based validation severely constrains the scalability of existing datasets. \Cref{fig:dataset_challenging} demonstrates the overview of the data collection process for quantum flake identification. 

\subsection{Limitations of Prior Dataset} 

Although MaskTerial~\cite{uslu2025maskterial} pioneered the use of physics-based rendering for generating large-scale 2D material datasets, its synthesis process still suffers from several limitations that reduce the realism and generalization of the generated data. First, the optical simulation is applied under idealized conditions without personalized color calibration, resulting in noticeable mismatches between the simulated and actual microscope images. Second, the previous method does not support the CIE color standard, which further limits the transferability of the dataset to diverse optical setups. These simplifications collectively widen the domain gap between synthetic and real data, constraining the practical usability of the generated dataset for robust model training. Lastly, prior work generates new flakes independently of the existing ones, which can result in unrealistic overlaps or physically inconsistent stacking between flakes in the same image.  \cref{tab:synthesis_comparison} demonstrates the pros and cons of the previous method compared to ours.

\section{Synthetic Materials Framework (Synthia)}

\subsection{Multilayer Optical Model}
\label{sec:tmm}

To synthesize realistic images of 2D material flakes on typical substrates, 
we employ multilayer optical modeling via the \textit{Transfer Matrix Method} (TMM)~\cite{byrnes2020multilayeropticalcalculations}. 
This physics-based model simulates thin-film interference effects, enabling wavelength-dependent prediction 
of reflectance and optical contrast as a function of flake thickness.

\noindent\textbf{Optical model.} 
Consider a multilayer stack with $L$ thin layers, each indexed by $l = 1, 2, \dots, L$. 
The incident and substrate media are denoted by layers $0$ and $L{+}1$, respectively, both assumed semi-infinite. 
Each layer $l$ is characterized by its complex refractive index 
$n_l(\lambda) + i k_l(\lambda)$ and physical thickness $d_l$. 
When light of wavelength $\lambda$ enters the stack at normal incidence, 
it undergoes multiple reflections and transmissions across layer interfaces. At each interface between layers $l$ and $l{+}1$, 
the reflection and transmission coefficients are determined by the Fresnel equations as in \cref{eq:fresnel}:
\begin{equation}
\label{eq:fresnel}
\footnotesize
r_{l,l+1} = \frac{n_l - n_{l+1}}{n_l + n_{l+1}}, 
\quad\quad
t_{l,l+1} = \frac{2n_l}{n_l + n_{l+1}}.
\end{equation}
Light propagation through each layer introduces a phase delay 
$\delta_l = \frac{2\pi n_l d_l}{\lambda}$, which can be expressed by a propagation matrix $P_l$. 
The total electric field across the multilayer stack can then be written using 
the transfer-matrix formalism as in \cref{eq:tmm_matrix}:
\begin{equation}
\footnotesize
\begin{pmatrix}
v \\[3pt] u
\end{pmatrix}
= M_{0,1}
\left( \prod_{l=1}^{L} P_l \, M_{l,l+1} \right)
\begin{pmatrix}
u_0 \\[3pt] 0
\end{pmatrix},
\label{eq:tmm_matrix}
\end{equation}
where $u_0=1$ is the incident field amplitude, 
$v$ is the reflected amplitude, and $u$ is the transmitted amplitude into the substrate. 
The reflectance is formed as
$R(\lambda) = \left| v / u_0 \right|^2$.

\noindent\textbf{Reflectance computation for synthetic flakes.}
In our synthetic pipeline, each flake is modeled as a multilayer stack consisting of: \textbf{(1)} air (incident medium), \textbf{(2)} the 2D material flake of thickness $d_\text{flake}$ corresponding to number of layers, \textbf{(3)} an oxide layer (e.g., SiO$_2$) of fixed thickness $d_\text{SiO2}$, and a silicon substrate (semi-infinite).
We sample wavelengths $\lambda \in [400, 700]$~nm at $D$ discrete intervals and compute $R(\lambda)$ 
using ~\cref{eq:tmm_matrix}.

\noindent\textbf{Color space conversion.}
To convert the simulated reflectance spectra into RGB values, 
we integrate the reflectance over the visible spectrum using 
the CIE 1931 color-matching functions $S(\lambda)$ and the standard illuminant spectrum $I(\lambda)$ as in \cref{eq:rgb_integral}.
\begin{equation}
\footnotesize
x = \int_{\lambda_\text{min}}^{\lambda_\text{max}} 
S(\lambda) \, I(\lambda) \, R(\lambda) \, d\lambda
\approx 
\sum_{i=1}^{D} S(\lambda_i) \, I(\lambda_i) \, R(\lambda_i),
\label{eq:rgb_integral}
\end{equation}
where $S(\lambda) \in \mathbb{R}^{3 \times 1}$ produces the $(R,G,B)$ components.
In matrix form, this can be expressed as $x = S^\top (I \circ R)$, 
where $\circ$ denotes the element-wise product.
\cref{fig:optical_model_data_comparison} (a) illustrates what the optical model looks like.

\noindent\textbf{Implementation.}
We implement the above optical model using the Python package \texttt{tmm}~\cite{byrnes2020multilayeropticalcalculations}, 
which provides efficient simulation of multilayer thin films. 
For each material (e.g., graphene, MoS$_2$, WSe$_2$, hBN), 
we define refractive index functions $n(\lambda)$ and $k(\lambda)$, 
and compute their reflectance spectra across thickness variations. 
The color-accurate RGB values are rendered as synthetic microscopy patches 
and combined with real microscope backgrounds.

\subsection{Details of Synthia}
\begin{figure*}[!t]
    \centering
    \includegraphics[width=0.8\linewidth]{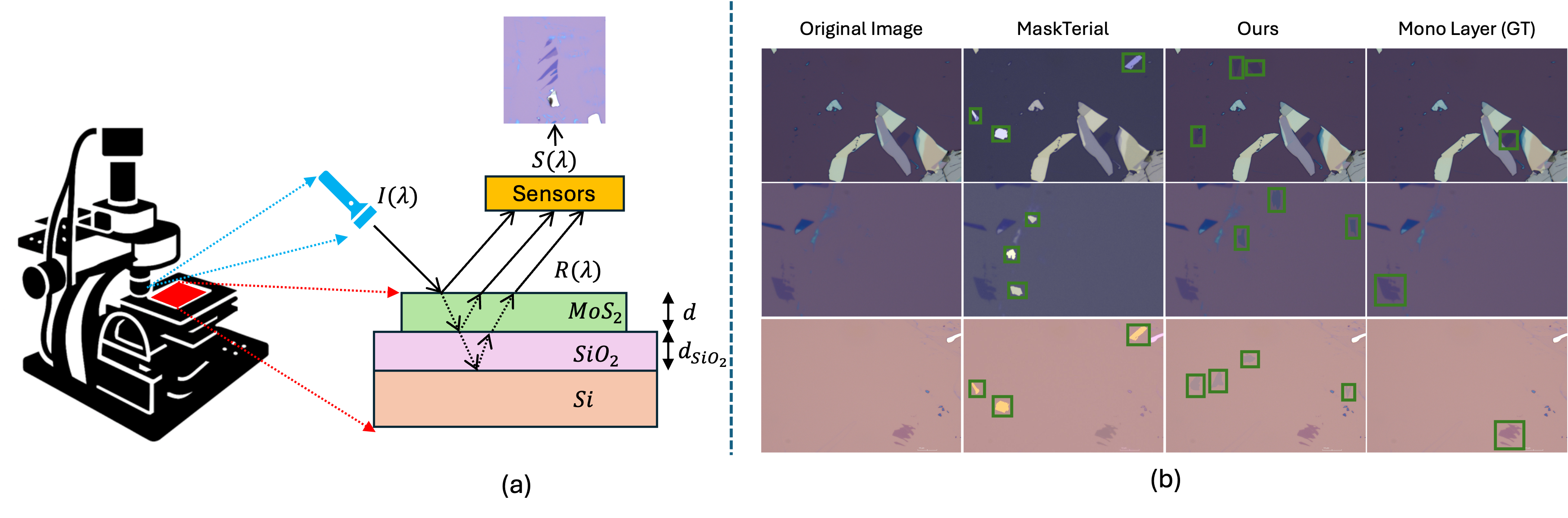}
    \vspace{-4mm}
    \caption{
    \textbf{(a)} Optical thin-film model used in \textbf{Synthia}, capturing layer-dependent interference between the material (e.g., MoS$_2$), SiO$_2$, and Si substrate under microscope illumination. 
    \textbf{(b)} Comparison of synthetic flake quality, showing that \textbf{Synthia} produces more realistic color contrast, edge appearance, and flake visibility across materials and thicknesses than previous approaches. \textbf{(Best view in colors)}}
    \label{fig:optical_model_data_comparison}
    \vspace{-4mm}
\end{figure*}

In this section, we introduce \textbf{Synthia}, a new synthetic 2D material generation framework that can produce a large number of realistic flakes under diverse conditions, including different material types, substrate kinds, and layer thicknesses. The core of Synthia is built upon the optical model described in \cref{sec:tmm}, which simulates thin-film interference to reproduce the color appearance of 2D materials. To further enhance realism and reduce the domain gap between synthetic and real data, we incorporate two additional components: the \textit{White-Balance-Aware} module and the \textit{Substrate-Aware} module. These modules account for illumination variability and substrate-dependent color shifts, respectively. Specifically, Synthia leverages a Physics-Informed Attention (PIA) module, denoted as $\mathcal{F}_{\text{PIA}}$, to identify physically meaningful substrate regions for flake placement. The details of the PIA module are presented later in \cref{sec:pia}. The complete pipeline of Synthia is detailed in \cref{alg:pgs}. \cref{fig:optical_model_data_comparison} (b) compares the realism of our synthesized flakes with the other method.

{
\vspace{-2mm}
\setlength{\textfloatsep}{0pt}
\begin{algorithm}[!b]
\caption{\textbf{Synthetic 2D Materials (Synthia)}}
\label{alg:pgs}
\footnotesize
\begin{algorithmic}[1]
\Require Reference microscopy image $\mathbf{I}_{\text{ref}}$, Optical model $\mathcal{T}$, 
Color projector $\Phi$, Material type $s$, Physics-Informed Attention function $\mathcal{F}_{\text{PIA}}$
\Ensure Synthetic microscopy image $\mathbf{I}_{\text{out}}$

\vspace{0.5em}
\Statex \textbf{// --- White-Balance-Aware Module ---}
\State $\mathbf{c}_{\text{sub}}^{\text{ref}} \gets \text{MedianColor}(\mathbf{I}_{\text{ref}})$
\State $R_{\text{sub}}(\lambda) \gets \mathcal{T}(\{\text{air}, \text{SiO}_2, \text{Si}\}, t_{\text{sub}}, \lambda)$
\State $\mathbf{c}_{\text{sub}}^{0} \gets \Phi(R_{\text{sub}}(\lambda), I(\lambda))$
\State $\mathbf{g} \gets \mathbf{c}_{\text{sub}}^{\text{ref}} \oslash \mathbf{c}_{\text{sub}}^{0}$ 
\Comment{Compute personalized white-balance gain based on reference substrate color}

\vspace{0.5em}
\Statex \textbf{// --- Substrate-Aware Module ---}
\State $\mathbf{A} \gets \mathcal{F}_{\text{PIA}}(\mathbf{I}_{\text{ref}})$
\State $\tilde{\mathbf{A}} \gets \text{Normalize}(\mathbf{A})$
\State $\mathbf{M}_{\text{sub}} \gets \mathbf{1}\{\tilde{\mathbf{A}} < \operatorname{Perc}_{90}(\tilde{\mathbf{A}})\}$ 
\Comment{Detect clean substrate areas for valid flake placement}

\vspace{0.5em}
\Statex \textbf{// --- Synthetic Flake Generation ---}
\For{$i = 1$ to $N_{\text{flakes}}$}
    \State $\mathbf{M}_i \gets \text{RandomFlakeMask()}$
    \State $t_i \gets \text{SampleThickness}(s)$
    \State Find valid coordinates $(u_i, v_i)$ such that 
           $\mathbf{M}_{\text{sub}}[v_i:v_i+h_i,\,u_i:u_i+w_i] \odot \mathbf{M}_i \equiv 0$
           \Comment{Avoid overlapping flakes}
    \State $R_i(\lambda) \gets \mathcal{T}(\{\text{air}, s, \text{SiO}_2, \text{Si}\}, t_i, \lambda)$
    \State $\mathbf{c}_i \gets \Phi(R_i(\lambda), I(\lambda))$ \Comment{Compute flake color via optical model}
    \State $\mathbf{c}_i \gets \mathbf{g} \odot \mathbf{c}_i$ 
           \Comment{Apply White-Balance-Aware correction}
    \State $\mathbf{I}_{\text{ref}}[v_i:v_i+h_i,\,u_i:u_i+w_i] 
           \gets \mathbf{c}_i \mathbf{M}_i 
           + \mathbf{I}_{\text{ref}}[v_i:v_i+h_i,\,u_i:u_i+w_i](1 - \mathbf{M}_i)$
\EndFor

\vspace{0.5em}
\State $\mathbf{I}_{\text{out}} \gets \mathbf{I}_{\text{ref}}$
\State \textbf{Return:} $\mathbf{I}_{\text{out}}$

\end{algorithmic}
\end{algorithm}
}

\subsection{QMat-Instruct Dataset}
\label{sec:qmat_instruct}

\noindent\textbf{Lack of a 2D dataset to train MLLM.} While previous studies in 2D material characterization mainly focused on image-based detection or segmentation, there is currently {no existing dataset designed for training MLLMs} to understand and reason about quantum flakes. 

\noindent\textbf{Proposed QMat-Instruct Dataset.} 
To support instruction tuning for quantum materials, we build \textbf{QMat-Instruct}, the first large-scale multimodal instruction dataset in this domain. Our data is enabled by realistic synthetic images from {Synthia}, allowing us to generate diverse and physics-consistent supervision at scale. Each sample is formatted as a structured question-answer pair that embeds \textbf{physics-aware descriptions} of 2D flakes, \eg, their contrast, color shift, and edge transparency. These cues come directly from thin-film interference and are essential for distinguishing mono-, bi-, and multi-layer flakes. Unlike prior methods that rely on hand-crafted rules or specialized models, QMat-Instruct teaches the model to learn the physical relationships {through both language and vision}. The instructions cover common tasks in quantum material characterization, including counting, localization, and binary verification. Examples include: ``How many monolayer flakes are in the image?'', ``Locate the monolayer flakes,'' and ``Does this sample contain a monolayer flake?''. Overall, QMat-Instruct provides the \textbf{first physics-informed instructional corpus for multimodal models in quantum materials}, offering rich supervision that helps the model understand and reason about flake appearance across materials and imaging setups.

\section{The Proposed Method}

\begin{figure}[!b]
    \centering
    \vspace{-6mm}
    \includegraphics[width=1\linewidth]{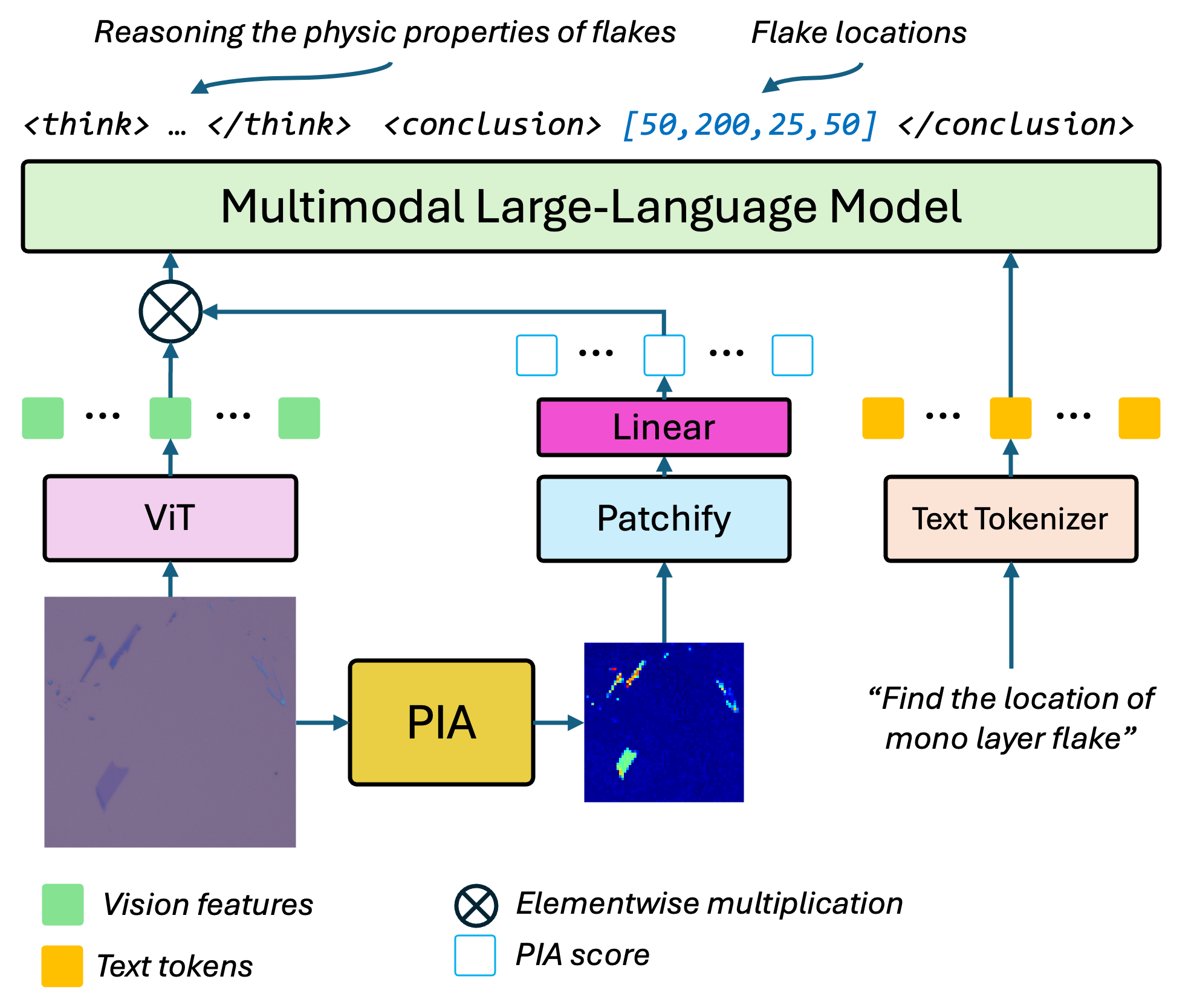}
    \vspace{-8mm}
    \caption{Overview of the proposed \METHOD framework. A Physics-Informed Attention (PIA) module extracts optical cues from the microscopy image, which are fused with ViT visual embeddings and text tokens in a multimodal large-language model for quantum material understanding.}
    \label{fig:architecture}
\end{figure}

\subsection{Physics-Aware Instruction Tuning (\METHOD)}
\label{subsec:method}
Our proposed \METHOD\ is illustrated in \cref{fig:architecture}.
Let $\mathbf{I} \in \mathbb{R}^{H \times W \times C}$ denote a microscopy image of a 2D material,
where $H$, $W$, and $C$ are the image height, width, and number of channels, respectively.
Let $\mathbf{T} = \{T_1, \dots, T_L\}$ denote a physics-informed instruction or question sequence
with $L$ tokens. 
The objective of our model is to enable a MLLM 
to reason about the optical properties of 2D flakes and generate physically grounded textual answers. We adopt a multimodal encoder–decoder architecture composed of 
(1) a Vision Transformer (ViT) guided by a Physics-Informed Attention (PIA) prior,
(2) a text encoder that embeds physics-aware instructions, and 
(3) an MLLM decoder for multimodal fusion and reasoning.

\vspace{0.3em}
\noindent
\textbf{Vision Encoder with Physics-Informed Attention.}
The vision encoder $\texttt{Enc}_v$ extracts $k$ patch embeddings from the input image as in \cref{eq:vision-encoder}:
\begin{equation} \label{eq:vision-encoder}
z_v = \texttt{Enc}_v(\mathbf{I}), 
\quad z_v = [\,\mathbf{v}_1, \dots, \mathbf{v}_k\,], 
\quad \mathbf{v}_i \in \mathbb{R}^d.
\end{equation}
The PIA module, introduced in \cref{sec:pia}, provides patch-level physics-based contrast scores 
$\boldsymbol{\alpha} = [\alpha_1, \dots, \alpha_k] \in [0,1]^k$ derived from optical cues such as thin-film interference and color contrast in CIELAB space. 
These scores reflect physical relevance but may be affected by illumination variability, white-balance bias, or sensor noise.

\noindent\textbf{Motivation.} While PIA encodes meaningful optical priors, it is a purely physics-driven process and cannot adapt to dataset-specific factors such as lighting, substrate color, or camera spectral response.
To enhance robustness while preserving physical interpretability, we introduce a \textbf{learnable linear correction layer} that adjusts the scale and bias of PIA scores.
This simple affine transformation enables the model to calibrate physics-based attention end-to-end through the language modeling loss, without requiring additional supervision.

Formally, the corrected patch attention weights are computed as in \cref{eq:weights_correction}:
\begin{equation}
\label{eq:weights_correction}
\beta_i = \sigma(w \alpha_i + b), \qquad
\beta_i \in [0,1], \quad w,b \in \mathbb{R},
\end{equation}
where $\sigma(\cdot)$ is the logistic sigmoid function.
The corrected weights $\beta_i$ are then used to modulate the visual tokens as in \cref{eq:visual_token_corrected}:
\begin{equation}
\label{eq:visual_token_corrected}
\tilde{\mathbf{v}}_i = \beta_i \cdot \mathbf{v}_i, 
\qquad
\tilde{z}_v = [\,\tilde{\mathbf{v}}_1, \dots, \tilde{\mathbf{v}}_k\,].
\end{equation}
This design retains the physically interpretable structure of PIA while allowing a small number of learnable parameters to adapt the attention distribution for better generalization.

\vspace{0.3em}
\noindent
\textbf{Text Encoder.}
The instruction $\mathbf{T}$ is tokenized and embedded using a pretrained text encoder as in \cref{eq:text-encoder}:
\begin{equation} \label{eq:text-encoder}
z_t = \texttt{Enc}_t(\mathbf{T}), 
\quad z_t = [\,\mathbf{t}_1, \dots, \mathbf{t}_L\,], 
\quad \mathbf{t}_j \in \mathbb{R}^d.
\end{equation}
These instructions incorporate physics-informed descriptions (e.g., “mono-layer flakes appear faint and semi-transparent”), which provide explicit physical guidance during multimodal training.

\vspace{0.3em}
\noindent
\textbf{Multimodal Fusion and Decoding.}
The physics-calibrated visual tokens $\tilde{z}_v$ and textual tokens $z_t$ are concatenated and input into an MLLM decoder $\texttt{LLM}$, which performs cross-modal attention and autoregressive text generation as in \cref{eq:llm}:
\begin{equation} \label{eq:llm}
\mathcal{Y} = \texttt{LLM}(\tilde{z}_v, z_t),
\quad
\mathcal{Y} = [\,y_1, \dots, y_M\,],
\end{equation}
where $\mathcal{Y}$ is the generated output sequence and $M$ is the number of tokens.

\vspace{0.3em}
\noindent
\textbf{Training Objective.}
The model is trained using a standard autoregressive loss on the physics-informed Q\&A corpus as in \cref{eq:loss-lm}:
\begin{equation} \label{eq:loss-lm}
\small
\mathcal{L}_{\text{LM}} =
- \frac{1}{M} \sum_{j=1}^{M}
\log P_\theta(y_j \mid y_{<j}, \tilde{z}_v, z_t).
\end{equation}
No explicit supervision is applied to PIA; the affine correction parameters $(w,b)$ are optimized implicitly through backpropagation from $\mathcal{L}_{\text{LM}}$, enabling end-to-end calibration of the physics prior.
During inference, given $(\mathbf{I}, \mathbf{T})$, the model outputs $\mathcal{Y}$ as physically grounded reasoning about flake count, thickness, and location.

\subsection{Physics-Informed Attention (PIA)}
\label{sec:pia}
\textbf{Motivations. }The optical contrast of two-dimensional (2D) materials observed under an optical microscope originates from thin-film interference between the flake and the SiO$_2$/Si substrate.
This phenomenon leads to characteristic color variations that correlate with the layer thickness and refractive index of the material.
However, modeling this behavior using a full optical transfer-matrix simulation requires precise knowledge of wavelength-dependent refractive indices, illumination spectra, and camera responses, which are often unavailable or vary across laboratories.
To overcome these limitations, we adopt the CIE~LAB color space as a perceptually uniform and device-independent representation that encodes color contrast similarly to human visual perception.
In the LAB space, the Euclidean distance $\Delta E$ between a flake region and its surrounding substrate provides a robust and illumination-invariant measure of perceptual contrast, which approximates the reflectance difference caused by thin-film interference.
Unlike RGB intensities that depend on hardware calibration and white balance, LAB values separate luminance ($L^*$) and chromatic components ($a^*$, $b^*$), enabling more stable discrimination of flakes with subtle optical contrast variations.
Therefore, the LAB representation offers a simple yet physically meaningful proxy to capture the qualitative optical characteristics of 2D materials without requiring full optical simulation.

\noindent\textbf{PIA Approximation of Optical Contrast.}
We propose a simple, physics-aware proxy to approximate the optical thin-film contrast of two-dimensional (2D) materials using CIELAB distances.
Let a pixel (or image patch) at position $x$ have spectral reflectance $R(x,\lambda)$ and the bare substrate (background) have reflectance $R_{\text{bg}}(\lambda)$.
Under illumination spectrum $E(\lambda)$ and camera sensitivities $\{S_c(\lambda)\}_{c\in\{R,G,B\}}$, the recorded RGB intensity is $I_c(x) = \int R(x,\lambda)\,E(\lambda)\,S_c(\lambda)\,d\lambda$ and $I^{\text{bg}}_c = \int R_{\text{bg}}(\lambda)\,E(\lambda)\,S_c(\lambda)\,d\lambda$, 
where $\mathbf{I}(x) = (I_R, I_G, I_B)$ and $\mathbf{I}_{\text{bg}} = (I^{\text{bg}}_R, I^{\text{bg}}_G, I^{\text{bg}}_B)$.
We transform the RGB values into CIELAB color space through the nonlinear mapping $\Phi: \mathbb{R}^3 \!\to\! \mathbb{R}^3$, i.e.,
$(L^*, a^*, b^*) = \Phi(\mathbf{I})$.
The perceptual color contrast (PIA score) at pixel $x$ is defined as in \cref{eqn:PIA-att}
\begin{equation}\label{eqn:PIA-att}
\footnotesize
\begin{split}
\Delta E(x) 
&= \big\| \Phi(\mathbf{I}(x)) - \Phi(\mathbf{I}_{\text{bg}}) \big\|_2 \\
&= \sqrt{(L^*(x) - L^*_{\text{bg}})^2 + (a^*(x) - a^*_{\text{bg}})^2 + (b^*(x) - b^*_{\text{bg}})^2}.
\end{split}
\raisetag{30pt}
\end{equation}
Proof of \cref{eqn:PIA-att} is in the appendix.
In practice, the background color $\mathbf{I}_{\text{bg}}$ is estimated as the median pixel value of the image (or a region of interest),
and $\Delta E(x)$ is computed over non-overlapping patches to form an attention map.

\begin{figure}[!b]
    \centering
    \vspace{-4mm}
    \includegraphics[width=0.99\linewidth]{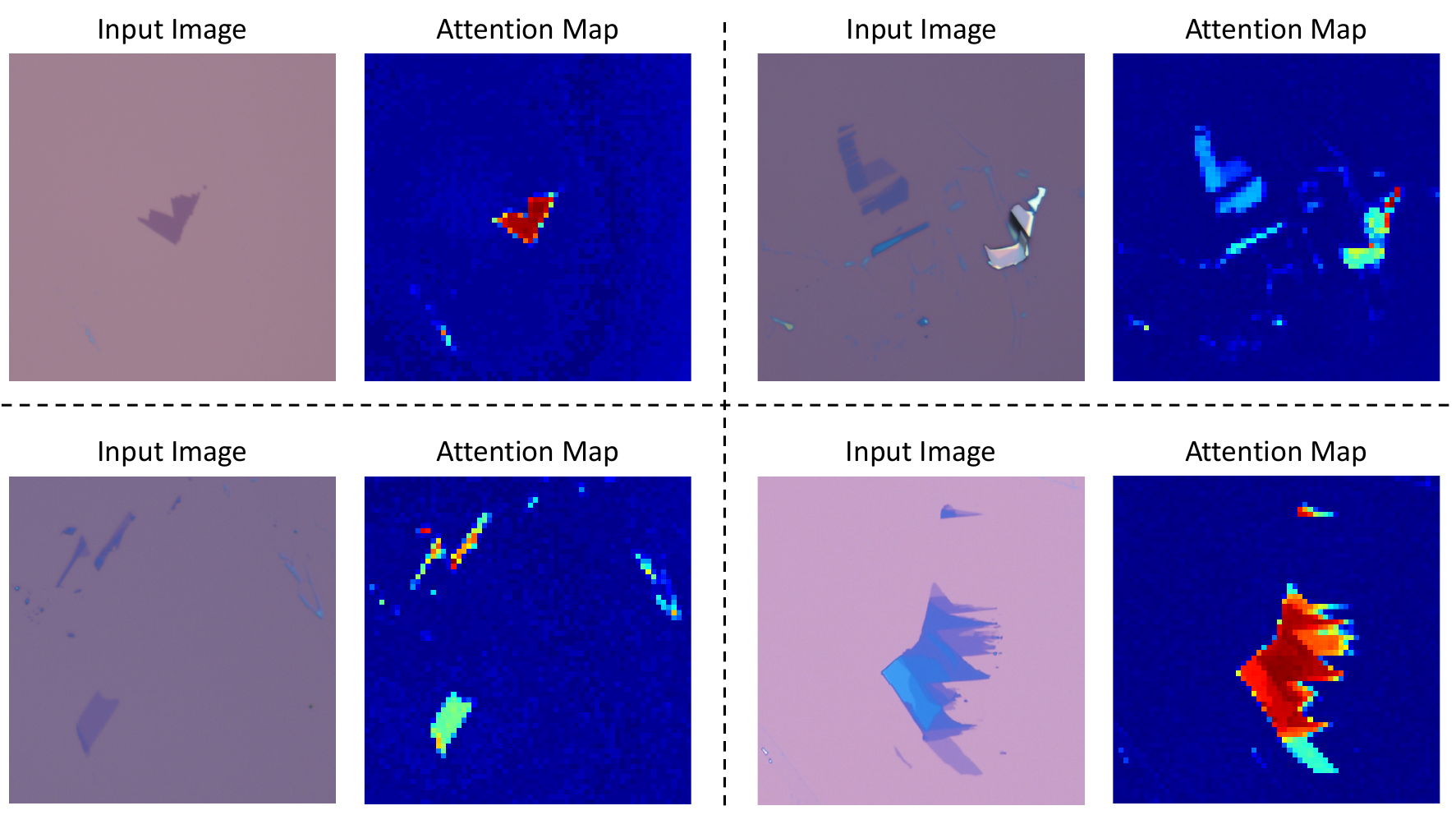}
    \vspace{-3mm}
    \caption{Physics-Informed Attention Map. \textbf{Best view in colors}.}
    \label{fig:pia_map}
\end{figure}

\noindent\textbf{Implementation.}
Given an image divided into patches $\{\mathcal{P}_k\}$, we compute $\mathbf{I}_{\text{bg}} = \operatorname{median}_{x \in \text{image}} \mathbf{I}(x), \widehat{\mathbf{I}}_k = \operatorname{median}_{x \in \mathcal{P}_k} \mathbf{I}(x)$
and define the patch-wise PIA scores as
\begin{equation}
\text{PIA-score}[k] = \big\| \Phi(\widehat{\mathbf{I}}_k) - \Phi(\mathbf{I}_{\text{bg}}) \big\|_2.
\end{equation}
This produces a stable and illumination-robust attention map that highlights 2D flakes through
their perceptual color contrast that is a direct image-space surrogate for the thin-film optical behavior
predicted by physical reflectance models. \cref{alg:lab_attention} illustrates the PIA module, and \cref{fig:pia_map} shows an example of the attention map.

{
\begin{algorithm}[t]
\caption{\textbf{Physics-Informed Attention Module}}
\label{alg:lab_attention}
\footnotesize
\begin{algorithmic}[1]
\Require Image $\mathbf{I} \in \mathbb{R}^{H \times W \times 3}$,
image size $(H, W)$, patch size $(h, w)$
\Ensure Perceptual attention map $\mathbf{A}_{\text{LAB}} \in \mathbb{R}^{H \times W}$

\Function{PIA}{$\mathbf{I}, (H, W), (h, w)$}
    \State Resize $\mathbf{I}$ to $(H, W)$
    \State $\mathbf{I}_{\text{bg}} \gets \operatorname{median}_{x \in \mathbf{I}}(\mathbf{I}(x))$ %
    \State $\mathbf{I}_{\text{bg}}^{\text{LAB}} \gets \Phi(\mathbf{I}_{\text{bg}})$ %

    \State Initialize attention list $\mathcal{A} = [\,]$
    \For{each patch $\mathcal{P}_k$ in $\mathbf{I}$ of size $(h, w)$}
        \State $\widehat{\mathbf{I}}_k \gets \operatorname{median}_{x \in \mathcal{P}_k}(\mathbf{I}(x))$ %
        \State $\widehat{\mathbf{I}}_k^{\text{LAB}} \gets \Phi(\widehat{\mathbf{I}}_k)$ %
        \State $\Delta E_k = \big\| \widehat{\mathbf{I}}_k^{\text{LAB}} - \mathbf{I}_{\text{bg}}^{\text{LAB}} \big\|_2$
        \State Append $\Delta E_k$ to $\mathcal{A}$
    \EndFor

    \State Reshape $\mathcal{A}$ into $\mathbf{A}_{\text{LAB}}$ of size $(H, W)$
    \State Normalize $\mathbf{A}_{\text{LAB}} \gets (\mathbf{A}_{\text{LAB}} - \min)/(\max - \min)$
    \State \Return $\mathbf{A}_{\text{LAB}}$
\EndFunction
\end{algorithmic}
\end{algorithm}
}

\section{Experiments}

\subsection{Implementation Details and Benchmarking}

\noindent\textbf{Implementation Details. } 
Our framework adopts the implementation of InternViT.
Input microscopy images are resized to $448\times448$ and divided into non-overlapping $14\times14$ patches. The text encoder is initialized from  Qwen3 \cite{yang2025qwen3}. We use AdamW optimizer with a learning rate of $2\times10^{-5}$, weight decay $1\times10^{-4}$, and batch size $4$.
The model is trained for $2$ epochs on 16 A100 GPUs. Data augmentation includes random cropping, horizontal flipping, and color jitter in CIELAB space to enhance robustness against illumination and white-balance variations. During inference, the model generates answers in a temperature-controlled sampling manner ($T{=}0.2$), enabling physically consistent reasoning on 2D flake characterization.

\noindent\textbf{Training Dataset.}
We train our models on a hybrid dataset combining real microscopy images with physics-based synthetic samples. Real flakes are collected under varied illumination, magnification, and 180\,nm SiO$_2$ substrates, following setups similar to prior work~\cite{masubuchi2020deep}. Each image contains about 30 annotated boxes. To compensate for the scarcity of thin-layer flakes, we synthesize data using thin-film optical simulation and color rendering that matches real microscope appearance. Our data includes both general scenes and focused regions with single-layer flakes across BN, Graphene, MoS$_2$, MoSe$_2$, MoWSe$_2$, WS$_2$, WSe$_2$, and WTe$_2$. Each image is paired with several physics-informed instruction–answer pairs describing optical cues and layer contrast, enabling the MLLM to align visual features with physics-grounded reasoning. In total, we generate 50,000 samples per material which yields 400,000 samples in total.

\noindent\textbf{Benchmark Dataset.}
Prior methods are typically evaluated on private datasets restricted to a single microscope, material, or imaging setup, which limits generalization. To provide a fair and comprehensive benchmark, we introduce \textbf{QF-Bench}, constructed from public datasets and our in-house collection. All samples are manually curated with bounding boxes and layer labels spanning mono-layer, few-layer (2-5 layers), and thick (5+ layers) flakes. The benchmark covers eight common 2D materials and captures diverse substrates and imaging conditions. \cref{tab:test-data-by-layer-mat} summarizes its statistics, including \textbf{280K} annotated flakes. QF-Bench offers the first unified platform for evaluating flake characterization models under realistic variability.

\begin{table}[t]
\footnotesize
\caption{Breakdown of annotated flake counts for each material by layer (Mono, Few, Thick).}
\label{tab:test-data-by-layer-mat}
\vspace{-3mm}
\centering
\begin{tabular}{l r r r r}
\toprule
\textbf{Material} & \textbf{Mono} & \textbf{Few} & \textbf{Thick} & \textbf{Total} \\
\midrule
BN        & 13   & 111  & 10{,}100 & 10{,}224 \\
Graphene  & 1{,}856 & 2{,}081 & 4{,}270 & 8{,}207 \\
MoS$_2$   & 246  & 536  & 108{,}352 & 109{,}134 \\
MoSe$_2$  & 24   & 32   & 1{,}327 & 1{,}383 \\
MoWSe$_2$ & 9    & 35   & 293 & 337 \\
WS$_2$    & 43   & 5    & 2{,}144 & 2{,}192 \\
WSe$_2$   & 207  & 35   & 6{,}195 & 6{,}437 \\
WTe$_2$   & 148  & 582  & 141{,}882 & 142{,}612 \\
\midrule
\textbf{Total} & \textbf{2{,}546} & \textbf{3{,}417} & \textbf{274{,}563} & \textbf{280{,}526} \\
\bottomrule
\end{tabular}
\vspace{-6mm}
\end{table}

\subsection{Experimental Results}

\noindent
\textbf{General Flake Detection.} 
As in \cref{tab:flake_identification}, across all detectors, our method shows a consistent improvement in general flake detection. Traditional baselines, \eg, MaskRCNN and ViTDet, achieve modest performance, with AP values ranging from 17--20\%. YOLOv11 models provide stronger results, improving AP to around 25--30\%, but still struggle to capture fine-grained flake boundaries, especially at higher IoU thresholds. Similarly, Co--DETR and RT--DETRv2 perform around 22--28\%. A flake detection method, MaskTerial \cite{uslu2025maskterial}, results in AP around 24\%. Another physics--based flake detection approach  \cite{nguyen2025varphiadaptphysicsinformedadaptationlearning} provides improved AP around 30\%. In contrast, our models achieve substantial gains across all metrics. \METHOD-1B already reaches 36.9 AP, outperforming YOLOv8-x by a large margin. As the model scale increases from 1B to 8B, the performance continues to improve steadily, reaching 45.6 AP and 60.5 AP$^{50}$. These results demonstrate that our multimodal architecture is more effective at modeling complex visual cues and material textures compared to conventional 2D detectors. The improvement at AP$^{75}$ is especially strong, showing that our method produces more accurate and tighter localization, which is critical for identifying thin flakes.

\noindent
\textbf{Mono Flake Detection.} 
As in \cref{tab:flake_identification}, mono-layer flake detection presents a more challenging setting due to their extremely thin structures and low contrast. All baseline detectors show a significant performance drop when moving from general flake detection to mono-layer detection. For example, YOLOv8-x drops from 28.3 AP to only 19.0 AP, confirming that single-layer flakes are difficult to localize using standard detection. MaskRCNN and ViTDet perform even worse, falling below 12 AP. Meanwhile, our method maintains strong performance and shows a much smaller performance drop. \METHOD-1B achieves 28.0 AP, already higher than all baseline models, and scaling up to \METHOD-8B further boosts accuracy to 37.3 AP and 52.8 AP$^{50}$. This shows that our model captures the subtle optical variations and boundary cues specific to mono-layer flakes. The improvements at AP$^{75}$ confirm sharper spatial reasoning, indicating that our approach is better at distinguishing true mono flakes from visually similar multi-layer regions.

\noindent \textbf{Instruction Grounding Evaluation.} We produce an instruction grounding evaluation. Specifically, we follow the standard protocols described in RefCOCO \cite{kazemzadeh2014referitgame}. As described in \Cref{tab:instruction_grounding}, we compare QuPAINT-2B against the fine-tuned versions of InternVL-2B and Qwen3VL-2B. Overall, given a comparable number of parameters, our approach outperforms the others on Acc@50 and Acc@75 due to the PIA module.

\begin{table}
\centering
\caption{Instruction Grounding Performance.}
\vspace{-3mm}
\resizebox{0.5\linewidth}{!}{
\begin{tabular}{ccc}
\hline
Method      & Acc@50 & Acc@75 \\
\hline
InternVL-2B & 43.3   & 30.8   \\
Qwen3VL-2B  & 45.8   & 32.1   \\
QuPAINT-2B  & 51.4   & 39.5  \\
\midrule
\end{tabular}
}
\label{tab:instruction_grounding}
\vspace{-6mm}
\end{table}

\vspace{-1mm}
\subsection{Ablation Studies}
\label{sec:ablation}

To understand the contributions of our approach, we conduct ablation studies focusing on two key factors: Physics-Aware Description (\cref{sec:qmat_instruct}) and Physical Interaction Attention (\cref{sec:pia}) on the mono flake detection task.

\noindent\textbf{Effect of PAD.}
As in \cref{tab:pir_pia_ablation}, the PAD module provides a physics-based prior that helps the model capture optical and material properties of thin flakes. Without PAD, the model obtains only 28.1 AP. Adding PAD alone significantly improves performance by $+3.6$ AP.
This improvement indicates that PAD helps the network learn more reliable representations and reduces confusion between mono-layer and multi-layer regions. PAD also improves localization quality, as shown by the higher AP$^{50}$ and AP$^{75}$ scores.

\noindent\textbf{Effect of PIA.}
As in \cref{tab:pir_pia_ablation}, the PIA module focuses the model on critical interaction patterns between the substrate, flake thickness, and lighting conditions. When used alone, PIA increases AP from 28.1 to 30.6. This gain demonstrates that PIA allows the network to attend to subtle cues that standard detectors tend to ignore. PIA strengthens mid- and high-IoU accuracy, leading to better spatial precision.

\noindent\textbf{Combined Effect of PAD and PIA.}
The full model, which integrates both PAD and PIA, achieves the best performance with 34.1 AP, 50.2 AP$^{50}$, and 38.0 AP$^{75}$ (\cref{tab:pir_pia_ablation}). These represent consistent improvements across all metrics. The combined gains show that PAD and PIA are complementary. PAD improves the feature representation from a physics standpoint, while PIA helps the model attend to the correct visual cues. Together, they enable more accurate and stable mono flake detection, especially in cases with low contrast and fine-scale boundaries.

\begin{table}[!t]
\centering
\setlength{\tabcolsep}{10pt}
\caption{Performance Comparison on Flake Detection Tasks.}
\vspace{-3mm}
\resizebox{\linewidth}{!}{
\begin{tabular}{l|ccc|ccc|cc}
\hline
\textbf{Detector}  & \multicolumn{3}{c|}{\textbf{General Flake Detection}} & \multicolumn{3}{c|}{\textbf{Mono Flake Detection}} & \textbf{\#Param} & \textbf{FLOPs} \\
 & \textbf{AP} & \textbf{AP$^{50}$} & \textbf{AP$^{75}$}
 & \textbf{AP} & \textbf{AP$^{50}$} & \textbf{AP$^{75}$} & \\
\hline
MaskRCNN-R50 \cite{he2017mask}       
& 18.7 & 36.5 & 17.8 
& 9.8 & 28.9 & 9.0 & 44M & 134.4B \\
MaskRCNN-R101 \cite{he2017mask}       
& 20.3 & 38.1 & 18.9 
& 11.1 & 30.6 & 10.3 & 63M & 334.8B \\
\hline
ViTDet (base) \cite{li2022exploringplainvisiontransformer}    
& 17.4 & 35.2 & 16.5 
& 8.7 & 27.6 & 7.8 & 86M & 1.3T \\
ViTDet (large) \cite{li2022exploringplainvisiontransformer} 
& 18.9 & 37.0 & 17.3 
& 9.9 & 29.8 & 8.5 & 300M & 4.1T \\
\hline
YOLOv11-m & 24.8 & 41.9 & 21.1 & 17.0 & 27.2 & 16.4 & 20M & 68.0B \\
YOLOv11-l & 26.9 & 43.0 & 23.5 & 18.9 & 29.8 & 17.3 & 25M & 86.9B \\
YOLOv11-x & 29.6 & 44.5 & 24.6 & 19.8 & 31.3 & 18.7 & 57M & 194.9B \\
\hline
Co-DETR & 22.7 & 35.9 & 23.3 & 12.2 & 25.3 & 11.4 & 304M & 119.4B \\
RT-DETRv2 & 27.5 & 37.2 & 27.0 & 17.5 & 37.8 & 14.7 & 76M & 259.0B \\
\hline
MaskTerial & 23.8 & 41.0 & 25.4 & 16.8 & 35.5 & 17.2 & 45M & 150.7B \\
$\varphi$-Adapt & 30.3 & 49.1 & 27.4 & 24.1 & 38.2 & 23.4 & 91M & 368.3B \\
\hline
{\METHOD-1B} 
& {36.9} & {50.7} & {38.6} 
& {28.0} & {42.9} & {29.5} & 1.1B & 1.5T \\
{\METHOD-2B} 
& {38.7} & {53.6} & {41.2} 
& {30.2} & {45.5} & {32.0} & 2.3B & 3.4T \\
{\METHOD-4B} 
& {42.4} & {58.2} & {46.8} 
& {34.1} & {50.2} & {38.0} & 4.7B & 7.9T \\
\textbf{\METHOD-8B} 
& \textbf{45.6} & \textbf{60.5} & \textbf{47.9} 
& \textbf{37.3} & \textbf{52.8} & \textbf{39.9} & \textbf{8.5B} & \textbf{14.5T} \\
\hline
\end{tabular}
}
\label{tab:flake_identification}
\end{table}

\begin{table}[!t]
\footnotesize
\centering
\setlength{\tabcolsep}{16pt}
\vspace{-4mm}
\caption{Ablation study on Physics-Informed Reasoning (PAD) and PIA modules for Mono Flake Detection.}
\vspace{-3mm}
\begin{tabular}{c|c|ccc}
\hline
\textbf{PAD} & \textbf{PIA} & \textbf{AP} & \textbf{AP$^{50}$} & \textbf{AP$^{75}$} \\
\hline
\xmark & \xmark & 28.1 & 42.7 & 30.3 \\
\xmark & \cmark  & 30.6 & 44.8 & 32.6 \\
\cmark  & \xmark & 31.7 & 46.5 & 33.4 \\
\cmark  & \cmark  & \textbf{34.1} & \textbf{50.2} & \textbf{38.0} \\
\hline
\end{tabular}
\label{tab:pir_pia_ablation}
\end{table}

\section{Conclusion and Limitations}
\noindent\textbf{Conclusions. } Our paper introduced a complete pipeline for quantum material characterization that combines physics-based data generation, instruction tuning, and multimodal reasoning. First, we proposed {Synthia}, a physics-grounded synthetic engine that simulates realistic optical responses of 2D materials. Second, we developed {QMat-Instruct}, a large-scale instruction dataset designed for multimodal models in quantum materials. Third, we presented \METHOD, a Physics-Aware Instruction Tuning framework that integrates a Physics-Informed Attention to improve flake representation and thickness recognition. Finally, we establish a new benchmark covering multiple materials, substrates, and imaging settings for fair and reproducible evaluation. 

\noindent\textbf{Limitations. } Our approach still has a few limitations. 
First, while Synthia produces realistic physics-based images, it cannot fully match all variations seen in real microscopes or all types of flake defects. As a result, some domain gaps may remain when applying the model to new labs or imaging conditions. 
Second, \METHOD\ is limited by the maximum token length of current multimodal models. When an image contains too many flakes, the number of visual tokens can exceed this limit, making it difficult for \METHOD\ to handle extremely dense scenes.

\section{Acknowledgment}
This work is partly supported by MonArk NSF Quantum Foundry, supported by the National Science Foundation QAMASE- i program under NSF award No. DMR-1906383. It acknowledges the Arkansas High-Performance Computing Center for providing GPUs.

{
    \small
    \bibliographystyle{ieeenat_fullname}
    \bibliography{main}

\begin{thebibliography}{49}
\providecommand{\natexlab}[1]{#1}
\providecommand{\url}[1]{\texttt{#1}}
\expandafter\ifx\csname urlstyle\endcsname\relax
  \providecommand{\doi}[1]{doi: #1}\else
  \providecommand{\doi}{doi: \begingroup \urlstyle{rm}\Url}\fi

\bibitem[Achiam et~al.(2023)Achiam, Adler, Agarwal, Ahmad, Akkaya, Aleman, Almeida, Altenschmidt, Altman, Anadkat, et~al.]{achiam2023gpt}
Josh Achiam, Steven Adler, Sandhini Agarwal, Lama Ahmad, Ilge Akkaya, Florencia~Leoni Aleman, Diogo Almeida, Janko Altenschmidt, Sam Altman, Shyamal Anadkat, et~al.
\newblock Gpt-4 technical report.
\newblock \emph{arXiv preprint arXiv:2303.08774}, 2023.

\bibitem[Assran et~al.(2025)Assran, Bardes, Fan, Garrido, Howes, Mojtaba, Komeili, Muckley, Rizvi, Roberts, Sinha, Zholus, Arnaud, Gejji, Martin, Hogan, Dugas, Bojanowski, Khalidov, Labatut, Massa, Szafraniec, Krishnakumar, Li, Ma, Chandar, Meier, LeCun, Rabbat, and Ballas]{assran2025vjepa2selfsupervisedvideo}
Mido Assran, Adrien Bardes, David Fan, Quentin Garrido, Russell Howes, Mojtaba, Komeili, Matthew Muckley, Ammar Rizvi, Claire Roberts, Koustuv Sinha, Artem Zholus, Sergio Arnaud, Abha Gejji, Ada Martin, Francois~Robert Hogan, Daniel Dugas, Piotr Bojanowski, Vasil Khalidov, Patrick Labatut, Francisco Massa, Marc Szafraniec, Kapil Krishnakumar, Yong Li, Xiaodong Ma, Sarath Chandar, Franziska Meier, Yann LeCun, Michael Rabbat, and Nicolas Ballas.
\newblock V-jepa 2: Self-supervised video models enable understanding, prediction and planning, 2025.

\bibitem[Bai et~al.(2023)Bai, Bai, Chu, Cui, Dang, Deng, Fan, Ge, Han, Huang, et~al.]{bai2023qwen}
Jinze Bai, Shuai Bai, Yunfei Chu, Zeyu Cui, Kai Dang, Xiaodong Deng, Yang Fan, Wenbin Ge, Yu Han, Fei Huang, et~al.
\newblock Qwen technical report.
\newblock \emph{arXiv preprint arXiv:2309.16609}, 2023.

\bibitem[Byrnes(2020)]{byrnes2020multilayeropticalcalculations}
Steven~J. Byrnes.
\newblock Multilayer optical calculations, 2020.

\bibitem[Chen et~al.(2024)Chen, Wu, Wang, Su, Chen, Xing, Zhong, Zhang, Zhu, Lu, et~al.]{chen2024internvl}
Zhe Chen, Jiannan Wu, Wenhai Wang, Weijie Su, Guo Chen, Sen Xing, Muyan Zhong, Qinglong Zhang, Xizhou Zhu, Lewei Lu, et~al.
\newblock Internvl: Scaling up vision foundation models and aligning for generic visual-linguistic tasks.
\newblock In \emph{Proceedings of the IEEE/CVF conference on computer vision and pattern recognition}, pages 24185--24198, 2024.

\bibitem[Courtney et~al.(2025)Courtney, Pendharkar, Bittner, Sharpe, and Goldhaber-Gordon]{courtney2025automated}
Elijah~DS Courtney, Mihir Pendharkar, Nathan~J Bittner, Aaron~L Sharpe, and David Goldhaber-Gordon.
\newblock Automated tabletop exfoliation and identification of monolayer graphene flakes.
\newblock \emph{Review of Scientific Instruments}, 96\penalty0 (5), 2025.

\bibitem[Crimmann et~al.(2025)Crimmann, Junker, Glauser, Lassaline, Nagamine, and Norris]{crimmann2025high}
Juri~G Crimmann, Moritz~N Junker, Yannik~M Glauser, Nolan Lassaline, Gabriel Nagamine, and David~J Norris.
\newblock High-throughput optical identification and statistical analysis of atomically thin semiconductors.
\newblock \emph{Advanced Optical Materials}, 13\penalty0 (16):\penalty0 2500150, 2025.

\bibitem[Dai et~al.(2023)Dai, Li, Li, Tiong, Zhao, Wang, Li, Fung, and Hoi]{dai2023instructblip}
Wenliang Dai, Junnan Li, Dongxu Li, Anthony Tiong, Junqi Zhao, Weisheng Wang, Boyang Li, Pascale Fung, and Steven Hoi.
\newblock Instruct{BLIP}: Towards general-purpose vision-language models with instruction tuning.
\newblock In \emph{Thirty-seventh Conference on Neural Information Processing Systems}, 2023.

\bibitem[Dendukuri et~al.(2020)Dendukuri, Keeling, Fereidouni, Burbridge, Luu, and Churchill]{dendukuri2020definingquantumneuralnetworks}
Aditya Dendukuri, Blake Keeling, Arash Fereidouni, Joshua Burbridge, Khoa Luu, and Hugh Churchill.
\newblock Defining quantum neural networks via quantum time evolution, 2020.

\bibitem[Dong et~al.(2025)Dong, Liu, Sun, Yang, Hu, Rao, and Liu]{dong2025insight}
Yuhao Dong, Zuyan Liu, Hai-Long Sun, Jingkang Yang, Winston Hu, Yongming Rao, and Ziwei Liu.
\newblock Insight-v: Exploring long-chain visual reasoning with multimodal large language models.
\newblock In \emph{Proceedings of the Computer Vision and Pattern Recognition Conference}, pages 9062--9072, 2025.

\bibitem[Han et~al.(2020)Han, Lin, Yang, Mao, Li, Wang, Yasuda, Wang, Fatemi, Zhou, et~al.]{han2020deep}
Bingnan Han, Yuxuan Lin, Yafang Yang, Nannan Mao, Wenyue Li, Haozhe Wang, Kenji Yasuda, Xirui Wang, Valla Fatemi, Lin Zhou, et~al.
\newblock Deep-learning-enabled fast optical identification and characterization of 2d materials.
\newblock \emph{Advanced Materials}, 32\penalty0 (29):\penalty0 2000953, 2020.

\bibitem[Hattori et~al.(2024)Hattori, Taniguchi, Watanabe, and Kitamura]{Hattori_2024}
Yoshiaki Hattori, Takashi Taniguchi, Kenji Watanabe, and Masatoshi Kitamura.
\newblock Identification of exfoliated monolayer hexagonal boron nitride films with a digital color camera under white light illumination.
\newblock \emph{Nanotechnology}, 35\penalty0 (37):\penalty0 375704, 2024.

\bibitem[He et~al.(2017)He, Gkioxari, Doll{\'a}r, and Girshick]{he2017mask}
Kaiming He, Georgia Gkioxari, Piotr Doll{\'a}r, and Ross Girshick.
\newblock Mask r-cnn.
\newblock In \emph{Proceedings of the IEEE international conference on computer vision}, pages 2961--2969, 2017.

\bibitem[Islam et~al.(2025)Islam, Khan, Mim, Rahman, Patwary, Islam, and Hossain]{ISLAM2025100161}
Md.~Aminul Islam, Safiullah Khan, Juhi~Jannat Mim, S~M~Maksudur Rahman, Md. Ahadul~Islam Patwary, Md.~Safiul Islam, and Nayem Hossain.
\newblock Recent advances of 2d materials in semiconductor application: A review.
\newblock \emph{Advanced Sensor and Energy Materials}, 4\penalty0 (4):\penalty0 100161, 2025.

\bibitem[J{\"a}ckering et~al.(2025)J{\"a}ckering, Wirth, Conrads, Profe, Rothstein, Kyoseva, Watanabe, Taniguchi, Kennes, Stampfer, et~al.]{jackering2025super}
Lina J{\"a}ckering, Konstantin~G Wirth, Lukas Conrads, Jonas~B Profe, Alexander Rothstein, Hristiyana Kyoseva, Kenji Watanabe, Takashi Taniguchi, Dante~M Kennes, Christoph Stampfer, et~al.
\newblock Super-resolution imaging of nanoscale inhomogeneities in hbn-covered and encapsulated few-layer graphene.
\newblock \emph{Advanced Science}, 12\penalty0 (14):\penalty0 2409039, 2025.

\bibitem[Jia et~al.(2021)Jia, Yang, Xia, Chen, Parekh, Pham, Le, Sung, Li, and Duerig]{jia2021scaling}
Chao Jia, Yinfei Yang, Ye Xia, Yi-Ting Chen, Zarana Parekh, Hieu Pham, Quoc Le, Yun-Hsuan Sung, Zhen Li, and Tom Duerig.
\newblock Scaling up visual and vision-language representation learning with noisy text supervision.
\newblock In \emph{International conference on machine learning}, pages 4904--4916. PMLR, 2021.

\bibitem[Kazemzadeh et~al.(2014)Kazemzadeh, Ordonez, Matten, and Berg]{kazemzadeh2014referitgame}
Sahar Kazemzadeh, Vicente Ordonez, Mark Matten, and Tamara Berg.
\newblock {R}efer{I}t{G}ame: Referring to objects in photographs of natural scenes.
\newblock In \emph{Proceedings of the 2014 Conference on Empirical Methods in Natural Language Processing ({EMNLP})}, pages 787--798, Doha, Qatar, 2014. Association for Computational Linguistics.

\bibitem[Kirillov et~al.(2023)Kirillov, Mintun, Ravi, Mao, Rolland, Gustafson, Xiao, Whitehead, Berg, Lo, et~al.]{kirillov2023segment}
Alexander Kirillov, Eric Mintun, Nikhila Ravi, Hanzi Mao, Chloe Rolland, Laura Gustafson, Tete Xiao, Spencer Whitehead, Alexander~C Berg, Wan-Yen Lo, et~al.
\newblock Segment anything.
\newblock In \emph{Proceedings of the IEEE/CVF international conference on computer vision}, pages 4015--4026, 2023.

\bibitem[Li et~al.(2023)Li, Li, Savarese, and Hoi]{li2023blip}
Junnan Li, Dongxu Li, Silvio Savarese, and Steven Hoi.
\newblock Blip-2: Bootstrapping language-image pre-training with frozen image encoders and large language models.
\newblock In \emph{International conference on machine learning}, pages 19730--19742. PMLR, 2023.

\bibitem[Li et~al.(2022)Li, Mao, Girshick, and He]{li2022exploringplainvisiontransformer}
Yanghao Li, Hanzi Mao, Ross Girshick, and Kaiming He.
\newblock Exploring plain vision transformer backbones for object detection, 2022.

\bibitem[Liu et~al.(2023)Liu, Li, Wu, and Lee]{liu2023visual}
Haotian Liu, Chunyuan Li, Qingyang Wu, and Yong~Jae Lee.
\newblock Visual instruction tuning.
\newblock \emph{Advances in neural information processing systems}, 36:\penalty0 34892--34916, 2023.

\bibitem[Liu et~al.(2025)Liu, Zhao, Xu, and Ghanem]{liu2025bolt}
Shuming Liu, Chen Zhao, Tianqi Xu, and Bernard Ghanem.
\newblock Bolt: Boost large vision-language model without training for long-form video understanding.
\newblock In \emph{Proceedings of the Computer Vision and Pattern Recognition Conference}, pages 3318--3327, 2025.

\bibitem[Luu et~al.(2024)Luu, NGUYEN, and Churchill]{luu2024automatically}
Khoa Luu, Xuan~Bac NGUYEN, and Hugh Churchill.
\newblock Automatically detecting false negative objects in 2d material detection data sets, 2024.
\newblock US Patent App. 18/443,058.

\bibitem[Masubuchi et~al.(2020)Masubuchi, Watanabe, Seo, Okazaki, Sasagawa, Watanabe, Taniguchi, and Machida]{masubuchi2020deep}
Satoru Masubuchi, Eisuke Watanabe, Yuta Seo, Shota Okazaki, Takao Sasagawa, Kenji Watanabe, Takashi Taniguchi, and Tomoki Machida.
\newblock Deep-learning-based image segmentation integrated with optical microscopy for automatically searching for two-dimensional materials.
\newblock \emph{npj 2D Materials and Applications}, 4\penalty0 (1):\penalty0 3, 2020.

\bibitem[McKenzie et~al.(2024)McKenzie, Sharma, and Liu]{mckenzie2024fabrication}
James McKenzie, Nileema Sharma, and Xiaolong Liu.
\newblock Fabrication of pristine 2d heterostructures for scanning probe microscopy.
\newblock \emph{APL Materials}, 12\penalty0 (7):\penalty0 070602, 2024.

\bibitem[Nguyen et~al.(2025)Nguyen, Nguyen, Pandey, Faltermeier, Borys, Churchill, and Luu]{nguyen2025varphiadaptphysicsinformedadaptationlearning}
Hoang-Quan Nguyen, Xuan~Bac Nguyen, Sankalp Pandey, Tim Faltermeier, Nicholas Borys, Hugh Churchill, and Khoa Luu.
\newblock $\varphi$-adapt: A physics-informed adaptation learning approach to 2d quantum material discovery, 2025.

\bibitem[Nguyen et~al.(2023)Nguyen, Churchill, Luu, and Khan]{nguyen2023quantum}
Xuan~Bac Nguyen, Hugh Churchill, Khoa Luu, and Samee~U Khan.
\newblock Quantum vision clustering.
\newblock \emph{arXiv preprint arXiv:2309.09907}, 2023.

\bibitem[Nguyen et~al.(2024)Nguyen, Bisht, Thompson, Churchill, Luu, and Khan]{nguyen2024two}
Xuan~Bac Nguyen, Apoorva Bisht, Benjamin Thompson, Hugh Churchill, Khoa Luu, and Samee~U Khan.
\newblock Two-dimensional quantum material identification via self-attention and soft-labeling in deep learning.
\newblock \emph{IEEE Access}, 12:\penalty0 139683--139691, 2024.

\bibitem[Ouaj et~al.(2023)Ouaj, Kramme, Metzelaars, Li, Watanabe, Taniguchi, Edgar, Beschoten, Kögerler, and Stampfer]{Ouaj_2023}
Taoufiq Ouaj, Leonard Kramme, Marvin Metzelaars, Jiahan Li, Kenji Watanabe, Takashi Taniguchi, James~H Edgar, Bernd Beschoten, Paul Kögerler, and Christoph Stampfer.
\newblock Chemically detaching hbn crystals grown at atmospheric pressure and high temperature for high-performance graphene devices.
\newblock \emph{Nanotechnology}, 34\penalty0 (47):\penalty0 475703, 2023.

\bibitem[Ouaj et~al.(2024)Ouaj, Arnold, Azpeitia, Baltic, Barjon, Cascales, Cun, Esteban, Garcia-Hernandez, Garnier, Gautam, Greber, Said~Hassani, Hemmi, Jiménez, Journet, Kögerler, Loiseau, Maestre, Metzelaars, Schmidt, Stampfer, Stenger, Steyer, Taniguchi, Toury, Watanabe, and Beschoten]{Ouaj_2024}
Taoufiq Ouaj, Christophe Arnold, Jon Azpeitia, Sunaja Baltic, Julien Barjon, José Cascales, Huanyao Cun, David Esteban, Mar Garcia-Hernandez, Vincent Garnier, Subodh~K Gautam, Thomas Greber, Said Said~Hassani, Adrian Hemmi, Ignacio Jiménez, Catherine Journet, Paul Kögerler, Annick Loiseau, Camille Maestre, Marvin Metzelaars, Philipp Schmidt, Christoph Stampfer, Ingrid Stenger, Philippe Steyer, Takashi Taniguchi, Bérangère Toury, Kenji Watanabe, and Bernd Beschoten.
\newblock Benchmarking the integration of hexagonal boron nitride crystals and thin films into graphene-based van der waals heterostructures.
\newblock \emph{2D Materials}, 12\penalty0 (1):\penalty0 015017, 2024.

\bibitem[Peng et~al.(2024)Peng, Li, Wu, Zhang, Sun, and Wu]{peng2024dfineredefineregressiontask}
Yansong Peng, Hebei Li, Peixi Wu, Yueyi Zhang, Xiaoyan Sun, and Feng Wu.
\newblock D-fine: Redefine regression task in detrs as fine-grained distribution refinement, 2024.

\bibitem[Pálinkás et~al.(2024)Pálinkás, Márity, Kandrai, Tajkov, Gmitra, Vancsó, Tapasztó, and Nemes-Incze]{PALINKAS2024119608}
András Pálinkás, Krisztián Márity, Konrád Kandrai, Zoltán Tajkov, Martin Gmitra, Péter Vancsó, Levente Tapasztó, and Péter Nemes-Incze.
\newblock Identification of graphite with perfect rhombohedral stacking by electronic raman scattering.
\newblock \emph{Carbon}, 230:\penalty0 119608, 2024.

\bibitem[Qian et~al.(2024)Qian, Li, Wu, Ye, Fei, Chua, Zhuang, and Tang]{qian2024momentor}
Long Qian, Juncheng Li, Yu Wu, Yaobo Ye, Hao Fei, Tat-Seng Chua, Yueting Zhuang, and Siliang Tang.
\newblock Momentor: Advancing video large language model with fine-grained temporal reasoning.
\newblock \emph{arXiv preprint arXiv:2402.11435}, 2024.

\bibitem[Radford et~al.(2021)Radford, Kim, Hallacy, Ramesh, Goh, Agarwal, Sastry, Askell, Mishkin, Clark, et~al.]{radford2021learning}
Alec Radford, Jong~Wook Kim, Chris Hallacy, Aditya Ramesh, Gabriel Goh, Sandhini Agarwal, Girish Sastry, Amanda Askell, Pamela Mishkin, Jack Clark, et~al.
\newblock Learning transferable visual models from natural language supervision.
\newblock In \emph{International conference on machine learning}, pages 8748--8763. PmLR, 2021.

\bibitem[Ravi et~al.(2024)Ravi, Gabeur, Hu, Hu, Ryali, Ma, Khedr, R{\"a}dle, Rolland, Gustafson, et~al.]{ravi2024sam}
Nikhila Ravi, Valentin Gabeur, Yuan-Ting Hu, Ronghang Hu, Chaitanya Ryali, Tengyu Ma, Haitham Khedr, Roman R{\"a}dle, Chloe Rolland, Laura Gustafson, et~al.
\newblock Sam 2: Segment anything in images and videos.
\newblock \emph{arXiv preprint arXiv:2408.00714}, 2024.

\bibitem[Redmon et~al.(2016)Redmon, Divvala, Girshick, and Farhadi]{redmon2016you}
Joseph Redmon, Santosh Divvala, Ross Girshick, and Ali Farhadi.
\newblock You only look once: Unified, real-time object detection.
\newblock In \emph{Proceedings of the IEEE conference on computer vision and pattern recognition}, pages 779--788, 2016.

\bibitem[Robinson et~al.(2025)Robinson, Robicheaux, Popov, Ramanan, and Peri]{robinson2025rfdetrneuralarchitecturesearch}
Isaac Robinson, Peter Robicheaux, Matvei Popov, Deva Ramanan, and Neehar Peri.
\newblock Rf-detr: Neural architecture search for real-time detection transformers, 2025.

\bibitem[Ronneberger et~al.(2015)Ronneberger, Fischer, and Brox]{ronneberger2015u}
Olaf Ronneberger, Philipp Fischer, and Thomas Brox.
\newblock U-net: Convolutional networks for biomedical image segmentation.
\newblock In \emph{International Conference on Medical image computing and computer-assisted intervention}, pages 234--241. Springer, 2015.

\bibitem[Siméoni et~al.(2025)Siméoni, Vo, Seitzer, Baldassarre, Oquab, Jose, Khalidov, Szafraniec, Yi, Ramamonjisoa, Massa, Haziza, Wehrstedt, Wang, Darcet, Moutakanni, Sentana, Roberts, Vedaldi, Tolan, Brandt, Couprie, Mairal, Jégou, Labatut, and Bojanowski]{siméoni2025dinov3}
Oriane Siméoni, Huy~V. Vo, Maximilian Seitzer, Federico Baldassarre, Maxime Oquab, Cijo Jose, Vasil Khalidov, Marc Szafraniec, Seungeun Yi, Michaël Ramamonjisoa, Francisco Massa, Daniel Haziza, Luca Wehrstedt, Jianyuan Wang, Timothée Darcet, Théo Moutakanni, Leonel Sentana, Claire Roberts, Andrea Vedaldi, Jamie Tolan, John Brandt, Camille Couprie, Julien Mairal, Hervé Jégou, Patrick Labatut, and Piotr Bojanowski.
\newblock Dinov3, 2025.

\bibitem[Steiner et~al.(2025)Steiner, Rahmel, Volmer, Windisch, Janssen, Pesch, Watanabe, Taniguchi, Libisch, Beschoten, et~al.]{steiner2025current}
Corinne Steiner, Rebecca Rahmel, Frank Volmer, Rika Windisch, Lars~H Janssen, Patricia Pesch, Kenji Watanabe, Takashi Taniguchi, Florian Libisch, Bernd Beschoten, et~al.
\newblock Current-induced brightening of vacancy-related emitters in hexagonal boron nitride.
\newblock \emph{Physical Review Research}, 7\penalty0 (3):\penalty0 L032037, 2025.

\bibitem[Tan et~al.(2025)Tan, Cao, Zhan, Xue, and Ding]{tan2025beyond}
Wentao Tan, Qiong Cao, Yibing Zhan, Chao Xue, and Changxing Ding.
\newblock Beyond human data: Aligning multimodal large language models by iterative self-evolution.
\newblock In \emph{Proceedings of the AAAI Conference on Artificial Intelligence}, pages 7202--7210, 2025.

\bibitem[Uslu et~al.(2024)Uslu, Ouaj, Tebbe, Nekrasov, Bertram, Sch{\"u}tte, Watanabe, Taniguchi, Beschoten, Waldecker, et~al.]{uslu2024open}
Jan-Lucas Uslu, Taoufiq Ouaj, David Tebbe, Alexey Nekrasov, Jo~Henri Bertram, Marc Sch{\"u}tte, Kenji Watanabe, Takashi Taniguchi, Bernd Beschoten, Lutz Waldecker, et~al.
\newblock An open-source robust machine learning platform for real-time detection and classification of 2d material flakes.
\newblock \emph{Machine Learning: Science and Technology}, 5\penalty0 (1):\penalty0 015027, 2024.

\bibitem[Uslu et~al.(2025)Uslu, Nekrasov, Hermans, Beschoten, Leibe, Waldecker, and Stampfer]{uslu2025maskterial}
Jan-Lucas Uslu, Alexey Nekrasov, Alexander Hermans, Bernd Beschoten, Bastian Leibe, Lutz Waldecker, and Christoph Stampfer.
\newblock Maskterial: A foundation model for automated 2d material flake detection.
\newblock \emph{Digital Discovery}, 2025.

\bibitem[Wegerhoff et~al.(2025)Wegerhoff, Scharfst{\"a}dt, Linden, and Bergschneider]{wegerhoff2025coherent}
Max Wegerhoff, Moritz Scharfst{\"a}dt, Stefan Linden, and Andrea Bergschneider.
\newblock Coherent interaction of 2 s and 1 s exciton states in transition-metal dichalcogenide monolayers.
\newblock \emph{Physical Review Letters}, 134\penalty0 (23):\penalty0 236901, 2025.

\bibitem[Xu et~al.(2025)Xu, Jin, Wu, Li, Song, Sun, and Yuan]{xu2025llava}
Guowei Xu, Peng Jin, Ziang Wu, Hao Li, Yibing Song, Lichao Sun, and Li Yuan.
\newblock Llava-cot: Let vision language models reason step-by-step.
\newblock In \emph{Proceedings of the IEEE/CVF International Conference on Computer Vision}, pages 2087--2098, 2025.

\bibitem[Yan et~al.(2025)Yan, Li, He, Wang, Li, Li, Zeng, Wang, Wang, Qiao, et~al.]{yan2025task}
Ziang Yan, Zhilin Li, Yinan He, Chenting Wang, Kunchang Li, Xinhao Li, Xiangyu Zeng, Zilei Wang, Yali Wang, Yu Qiao, et~al.
\newblock Task preference optimization: Improving multimodal large language models with vision task alignment.
\newblock In \emph{Proceedings of the Computer Vision and Pattern Recognition Conference}, pages 29880--29892, 2025.

\bibitem[Yang et~al.(2025)Yang, Li, Yang, Zhang, Hui, Zheng, Yu, Gao, Huang, Lv, et~al.]{yang2025qwen3}
An Yang, Anfeng Li, Baosong Yang, Beichen Zhang, Binyuan Hui, Bo Zheng, Bowen Yu, Chang Gao, Chengen Huang, Chenxu Lv, et~al.
\newblock Qwen3 technical report.
\newblock \emph{arXiv preprint arXiv:2505.09388}, 2025.

\bibitem[Zhang et~al.(2024)Zhang, Liu, Reid, Hartley, Zhuang, and Tang]{zhang2024motion}
Zeyu Zhang, Akide Liu, Ian Reid, Richard Hartley, Bohan Zhuang, and Hao Tang.
\newblock Motion mamba: Efficient and long sequence motion generation.
\newblock In \emph{European Conference on Computer Vision}, pages 265--282. Springer, 2024.

\bibitem[Zhou et~al.(2025)Zhou, Liu, Mo, Li, Peng, and Liu]{Zhou_2025_ICCV}
Hongliang Zhou, Yongxiang Liu, Canyu Mo, Weijie Li, Bowen Peng, and Li Liu.
\newblock When pixel difference patterns meet vit: Pidivit for few-shot object detection.
\newblock In \emph{Proceedings of the IEEE/CVF International Conference on Computer Vision (ICCV)}, pages 24309--24318, 2025.

\end{thebibliography}
}

\clearpage
\setcounter{page}{1}
\maketitlesupplementary

\section{Proof of \cref{eqn:PIA-att}}
\noindent\textbf{Proof (First-order Link to Optical Model).}
Let $\Delta R(\lambda) = R(x,\lambda) - R_{\text{bg}}(\lambda)$ denote the spectral difference caused by the 2D flake,
which depends on the refractive index $n(\lambda)$, thickness $d$, and interference phase $\phi = 2\pi n(\lambda)d / \lambda$.
Assuming $\|\Delta R\|_\infty$ is small (a valid approximation for mono- and few-layer flakes),
we can linearize $\Phi \circ \mathbf{I}$ about $\mathbf{I}_{\text{bg}}$ using a first-order Taylor expansion:
\begin{equation}
\small
\Delta E(x)
\approx 
\big\| J_{\Phi}(\mathbf{I}_{\text{bg}}) 
\underbrace{\int \Delta R(\lambda)\,E(\lambda)\,\mathbf{S}(\lambda)\,d\lambda}_{\Delta \mathbf{I}}
\big\|_2,
\end{equation}
where $J_{\Phi}$ is the Jacobian of the RGB-to-LAB transformation at $\mathbf{I}_{\text{bg}}$
and $\mathbf{S}(\lambda) = (S_R(\lambda), S_G(\lambda), S_B(\lambda))^{\top}$.
Therefore, there exists a positive constant $c$ (dependent on $E, \mathbf{S}, J_{\Phi}$) such that
\begin{equation}
\footnotesize
    \Delta E(x) \ge c \, \big\| \Delta \mathbf{I} \big\|_2  = c \, \Big\| \int \Delta R(\lambda)\,E(\lambda)\,\mathbf{S}(\lambda)\,d\lambda \Big\|_2.
\end{equation}
Since the thin-film interference model predicts that $\Delta R(\lambda; d)$
varies systematically with the number of layers $d$,
the perceptual contrast $\Delta E(x)$ is a monotonic proxy (locally)
for the reflectance deviation induced by optical interference.
Thus, maximizing $\Delta E$ over image patches identifies regions whose spectra differ from the substrate
due to interference, without explicitly solving the transfer-matrix model.

\section{More Details of Synthia}
In this section, we further discuss in detail the Synthia framework.
\subsection{White Balance Calibration}

After the reflected light is captured by the microscope, users often perform a manual color calibration process, typically by applying white balance correction to the image. This step adjusts the gain factors of the red, green, and blue (RGB) channels to compensate for color casts introduced by the illumination source or optical path. In practice, this calibration alters the perceived color of the substrate and flakes. From the perspective of material scientists, such calibration serves two main purposes: (1) it enhances the visibility of the \textit{target flakes}, making them more distinguishable from the background and other overlapping flakes; and (2) each user tends to develop personalized calibration preferences based on their experience and specific material systems, optimizing contrast for their observation workflow. As a result, microscopy images collected across different laboratories often exhibit noticeable color variations, even when using the same materials, substrates, and hardware configurations. This variability poses a significant challenge for AI models, which must remain robust to inconsistent color calibration settings. Therefore, in our synthetic pipeline, we explicitly model color calibration variability by simulating diverse white balance transformations. This step ensures that the generated synthetic flakes maintain realistic color contrast relative to their substrates and better reflect the diversity of real-world microscopy datasets.

\noindent\textbf{Implementation of White-Balance Calibration.}
Let the reference microscopy image be 
$\mathbf{I}_{\mathrm{ref}} \in \mathbb{R}^{3\times H\times W}$ 
in linear RGB space, captured for a known material and substrate configuration 
(e.g., MoS$_2$ on Si/SiO$_2$ with known oxide thickness). 
We denote by $\mathbf{M}_{\mathrm{sub}} \in \{0,1\}^{H\times W}$ 
the binary mask identifying substrate regions, and by $\mathbf{1}\{\cdot\}$ 
the indicator function. 
The mean substrate color after user calibration (white balance already applied) 
is computed as:
\begin{align}
[\mathbf{c}_{\mathrm{sub}}^{\mathrm{ref}}]_k
= \frac{\sum_{i,j} \mathbf{I}_{\mathrm{ref}}[k,i,j] \; 
\mathbf{M}_{\mathrm{sub}}[i,j]}{\sum_{i,j} \mathbf{M}_{\mathrm{sub}}[i,j]},
\quad k \in \{1,2,3\}.
\end{align}

Given the known optical stack (air / flake / SiO$_2$ / Si), 
the transfer-matrix module $\mathcal{T}$ computes the substrate reflectance spectrum 
$\mathbf{R}_{\mathrm{sub}}(\lambda) \in \mathbb{R}^{D}$ 
sampled across $D$ discrete wavelengths. 
A colorimetric projection function $\Phi$ (CIE 1931 color-matching with standard illuminant) 
maps spectra to linear RGB, yielding the pre-white-balance substrate color:
\begin{equation}
\mathbf{c}_{\mathrm{sub}}^{0} = 
\Phi\!\big(\mathbf{R}_{\mathrm{sub}}(\lambda)\big)
\in \mathbb{R}^3.
\end{equation}

Assuming a per-channel diagonal white balance model, 
the user’s gain vector $\mathbf{g} \in \mathbb{R}^3_{+}$ satisfies:
\begin{equation}
\mathbf{c}_{\mathrm{sub}}^{\mathrm{ref}} 
\approx \mathrm{diag}(\mathbf{g}) \; \mathbf{c}_{\mathrm{sub}}^{0}
\quad \Longrightarrow \quad
\mathbf{g} = 
\mathbf{c}_{\mathrm{sub}}^{\mathrm{ref}} \oslash \mathbf{c}_{\mathrm{sub}}^{0},
\end{equation}
where $\oslash$ denotes element-wise division 
(optionally normalized, e.g., $\frac{1}{3}\|\mathbf{g}\|_{1}=1$ to fix global exposure).

Finally, for any synthetic image before white balance 
$\mathbf{I}_{\mathrm{syn}}^{0} \in \mathbb{R}^{3\times h\times w}$, 
the personalized correction is applied as:
\begin{equation}
\mathbf{I}_{\mathrm{syn}}[k,i,j] 
= g_k \cdot \mathbf{I}_{\mathrm{syn}}^{0}[k,i,j],
\qquad k \in \{1,2,3\},
\end{equation}
followed by clipping to $[0,1]$. 
This process aligns the synthetic color distribution to the user’s 
individual calibration preferences while preserving the physically 
predicted flake–substrate contrast computed by $\mathcal{T}$ and $\Phi$.

\subsection{Substrate-Aware Synthetic Placement}
Previous data synthesis pipelines often overlook the spatial context of existing flakes in the reference microscopy images. As a result, newly generated synthetic flakes are randomly overlaid on the image, frequently overlapping with real flakes or debris already present on the substrate. Such unawareness introduces unrealistic artifacts and degrades the physical plausibility of the synthesized dataset. A straightforward solution would be to manually annotate bounding boxes for all visible flakes and restrict synthetic placement to the remaining background areas. However, this approach is highly labor-intensive, prone to human error, and does not scale to large microscopy collections. To overcome these limitations, we propose an automated \emph{substrate-detection} algorithm that localizes clean substrate regions directly from the input image. By identifying substrate-only areas through reflectance consistency and color uniformity, our method ensures that synthetic flakes are placed exclusively on physically valid regions, avoiding overlaps with existing structures. This substrate-aware synthesis greatly improves dataset realism and scalability without requiring manual annotations.

\subsection{Substrate-Aware Synthetic Flake Generation}
Given an unlabeled microscopy image 
$\mathbf{I}\in[0,255]^{H\times W\times3}$, 
our goal is to synthesize new flakes while avoiding overlap with existing ones. 
We first compute a LAB-based attention map 
$\mathbf{A} = \mathcal{F}_{\text{LAB}}(\mathbf{I}) \in \mathbb{R}^{H\times W}$ 
and normalize it to the range $[0,1]$:
\begin{equation}
\tilde{\mathbf{A}} = 
\frac{\mathbf{A} - \min(\mathbf{A})}
{\max(\mathbf{A}) - \min(\mathbf{A}) + \varepsilon}.
\end{equation}
High-attention regions correspond to existing flakes or high-texture areas. 
The substrate mask is then defined as
\begin{equation}
\mathbf{M}_{\text{sub}} = 
\mathbf{1}\!\left\{\tilde{\mathbf{A}} < \tau\right\}, 
\qquad
\tau = \operatorname{Perc}_{90}(\tilde{\mathbf{A}}),
\end{equation}
where $\tau$ denotes the 90th percentile threshold, 
preserving only low-attention (substrate) regions.

For each synthetic flake $f_i$ with binary mask $\mathbf{M}_i$ 
and sampled thickness $t_i$, 
we randomly select a valid substrate region $(u_i, v_i)$ that satisfies
\begin{equation}
\mathbf{M}_{\text{sub}}[v_i:v_i+h_i,\, u_i:u_i+w_i] 
\odot \mathbf{M}_i \equiv 0,
\end{equation}
ensuring no overlap with existing flakes or previously placed synthetic ones. 

\paragraph{Optical Color Generation.}
The color of each flake is physically computed using the 
\textit{transfer matrix method} (TMM). 
Given a material $s$ (e.g., MoS$_2$), 
a layer stack $\mathcal{S} = \{\text{air},\, s,\, \text{SiO}_2,\, \text{Si}\}$, 
and a flake thickness $t_i$, 
the wavelength-dependent reflectance spectrum is obtained by
\begin{equation}
R_i(\lambda) = \mathcal{T}(\mathcal{S}, t_i, \lambda),
\end{equation}
where $\mathcal{T}$ denotes the multilayer optical simulation. 
The resulting reflectance is converted into a perceptual RGB color via 
CIE 1931 color-matching functions $S(\lambda)$ and illuminant spectrum $I(\lambda)$:
\begin{equation}
\mathbf{c}_i = 
\Phi\!\big(R_i(\lambda), I(\lambda)\big)
= 
\int_{\lambda_{\min}}^{\lambda_{\max}}
S(\lambda)\, I(\lambda)\, R_i(\lambda)\, d\lambda,
\end{equation}
where $\mathbf{c}_i \in [0,255]^3$ represents the flake’s RGB color.

\paragraph{Flake Composition.}
The synthetic flake is placed directly on the selected substrate area by replacing 
the corresponding region in $\mathbf{I}$:
\begin{align}
&\mathbf{I}_{\text{out}}[v_i:v_i+h_i,\, u_i:u_i+w_i]
=  
\mathbf{c}_i\,\mathbf{M}_i \\ \nonumber
& \quad \quad \quad \quad + \mathbf{I}[v_i:v_i+h_i,\, u_i:u_i+w_i]\,(1 - \mathbf{M}_i),
\end{align}
where $\mathbf{M}_i\in\{0,1\}^{h_i\times w_i}$ denotes the binary flake mask.
This substrate-aware process ensures that synthetic flakes are placed 
only on physically valid substrate regions, 
avoiding overlaps and preserving realistic color consistency 
through the optical model.

\section{QMat-Instruct Dataset}
\begin{figure*}
    \centering
    \includegraphics[width=1\linewidth]{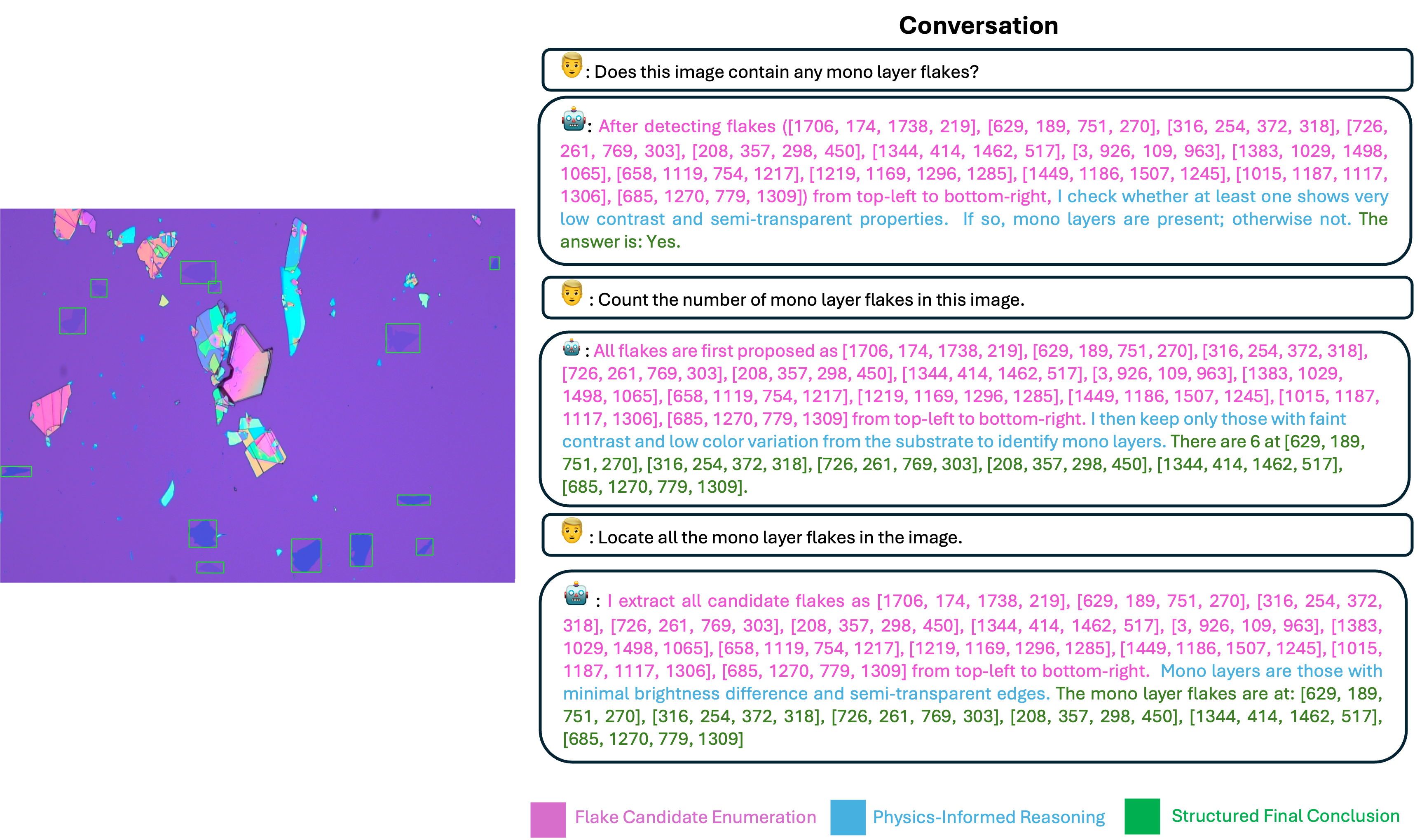}
    \caption{Conversation between a human and model analyzing a synthetic optical microscopy image. The model output is highlighted in three parts: flake candidate enumeration (pink), physics-informed reasoning (blue), and final conclusion (green).}
    \label{fig:synth_conversation1}
\end{figure*}

\begin{figure*}
    \centering
    \includegraphics[width=1\linewidth]{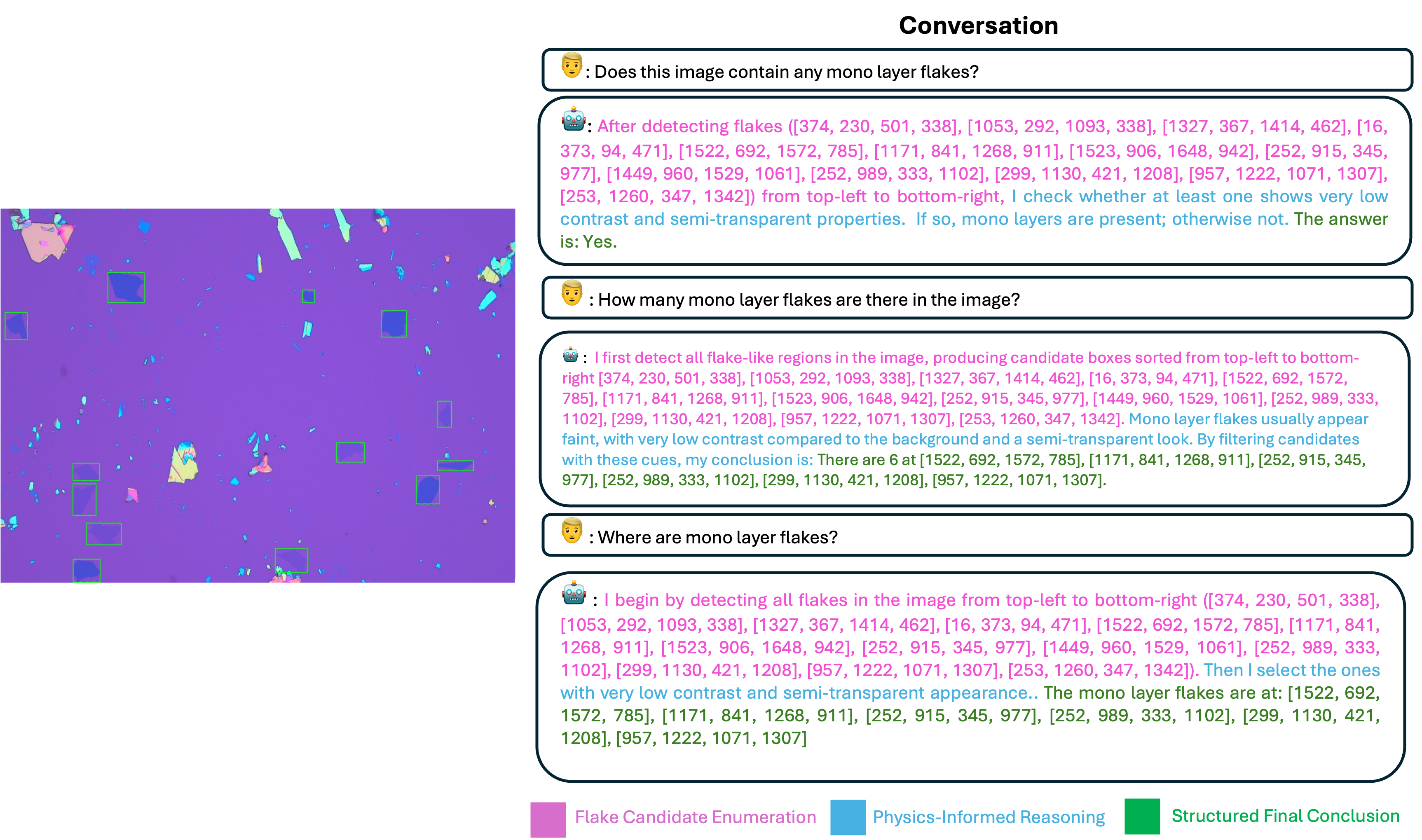}
    \caption{Conversation between a human and model analyzing a synthetic optical microscopy image. The model output is highlighted in three parts: flake candidate enumeration (pink), physics-informed reasoning (blue), and final conclusion (green).}
    \label{fig:synth_conversation2}
\end{figure*}

\begin{figure*}
    \centering
    \includegraphics[width=1\linewidth]{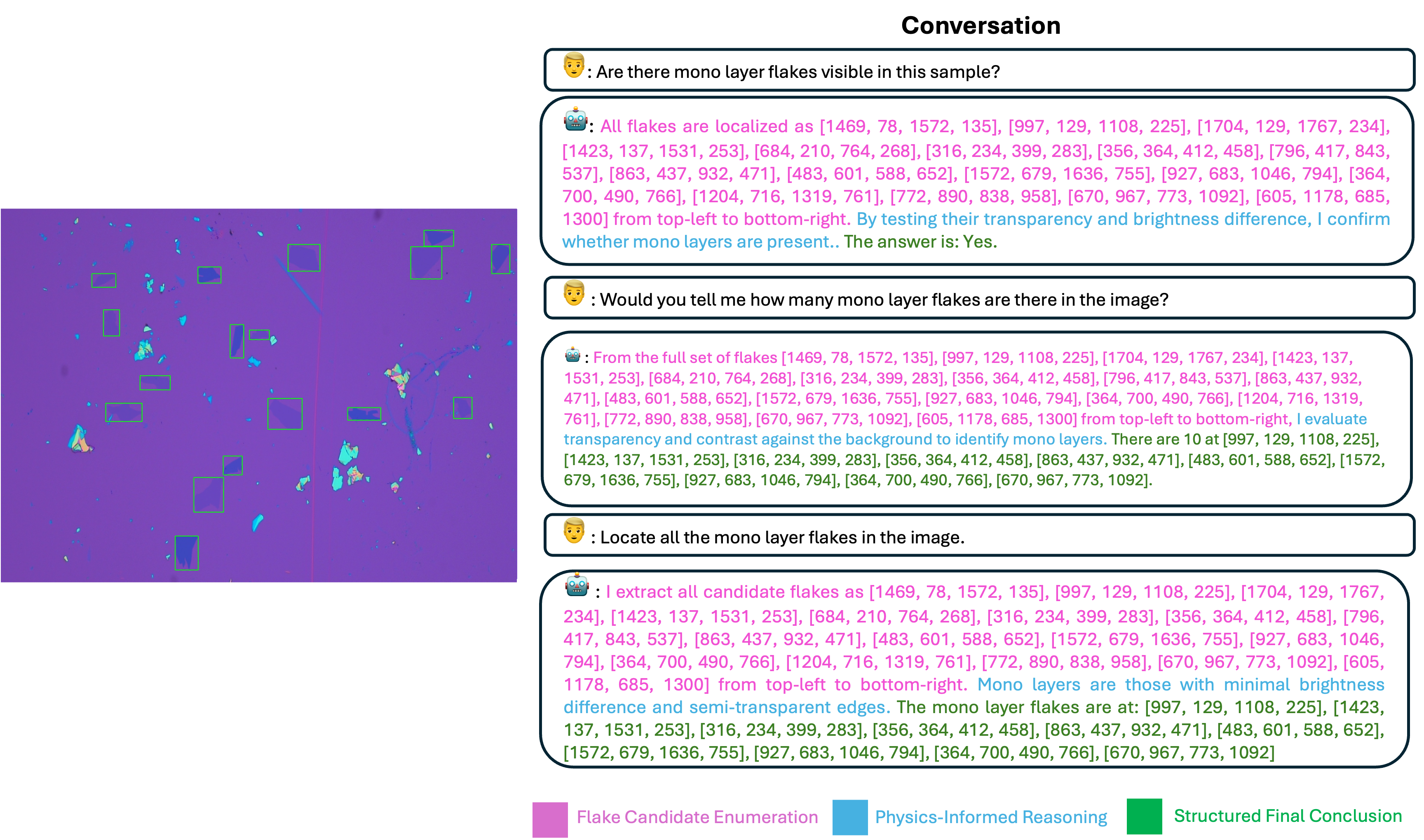}
    \caption{Conversation between a human and model analyzing a synthetic optical microscopy image. The model output is highlighted in three parts: flake candidate enumeration (pink), physics-informed reasoning (blue), and final conclusion (green).}
    \label{fig:synth_conversation3}
\end{figure*}

\begin{figure*}
    \centering
    \includegraphics[width=1\linewidth]{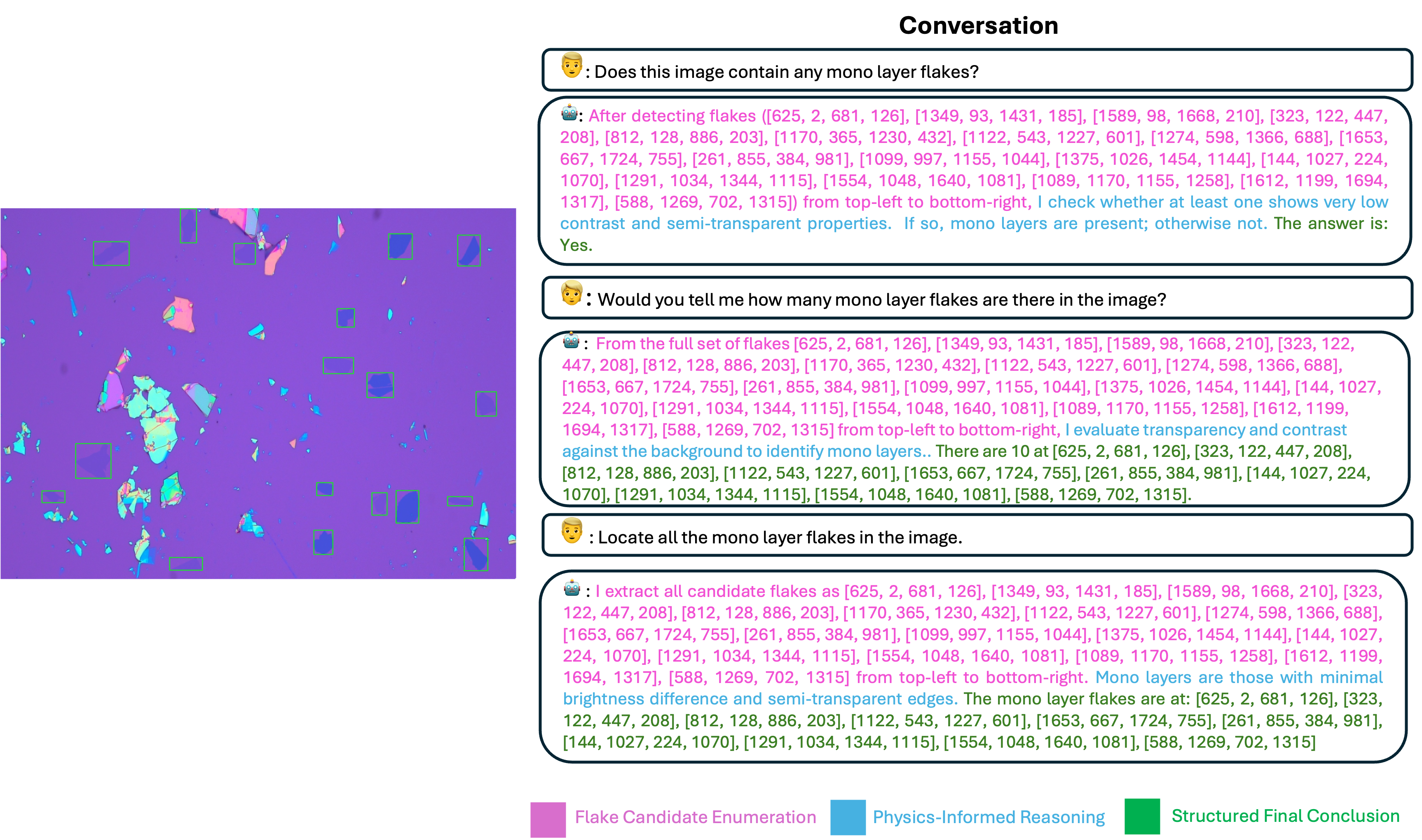}
    \caption{Conversation between a human and model analyzing a synthetic optical microscopy image. The model output is highlighted in three parts: flake candidate enumeration (pink), physics-informed reasoning (blue), and final conclusion (green).}
    \label{fig:synth_conversation4}
\end{figure*}

\begin{figure*}
    \centering
    \includegraphics[width=1\linewidth]{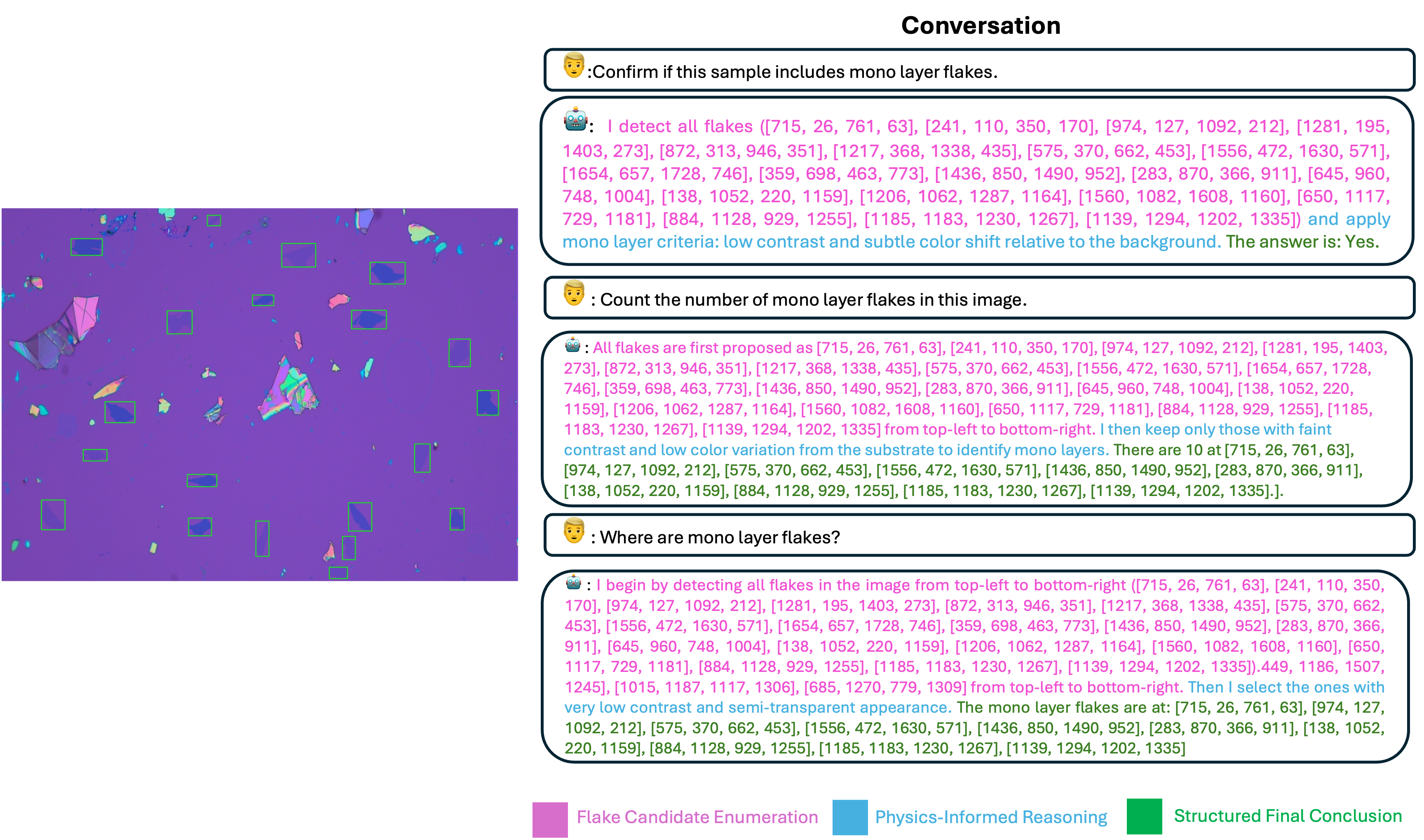}
    \caption{Conversation between a human and model analyzing a synthetic optical microscopy image. The model output is highlighted in three parts: flake candidate enumeration (pink), physics-informed reasoning (blue), and final conclusion (green).}
    \label{fig:synth_conversation5}
\end{figure*}

\begin{figure*}
    \centering
    \includegraphics[width=1\linewidth]{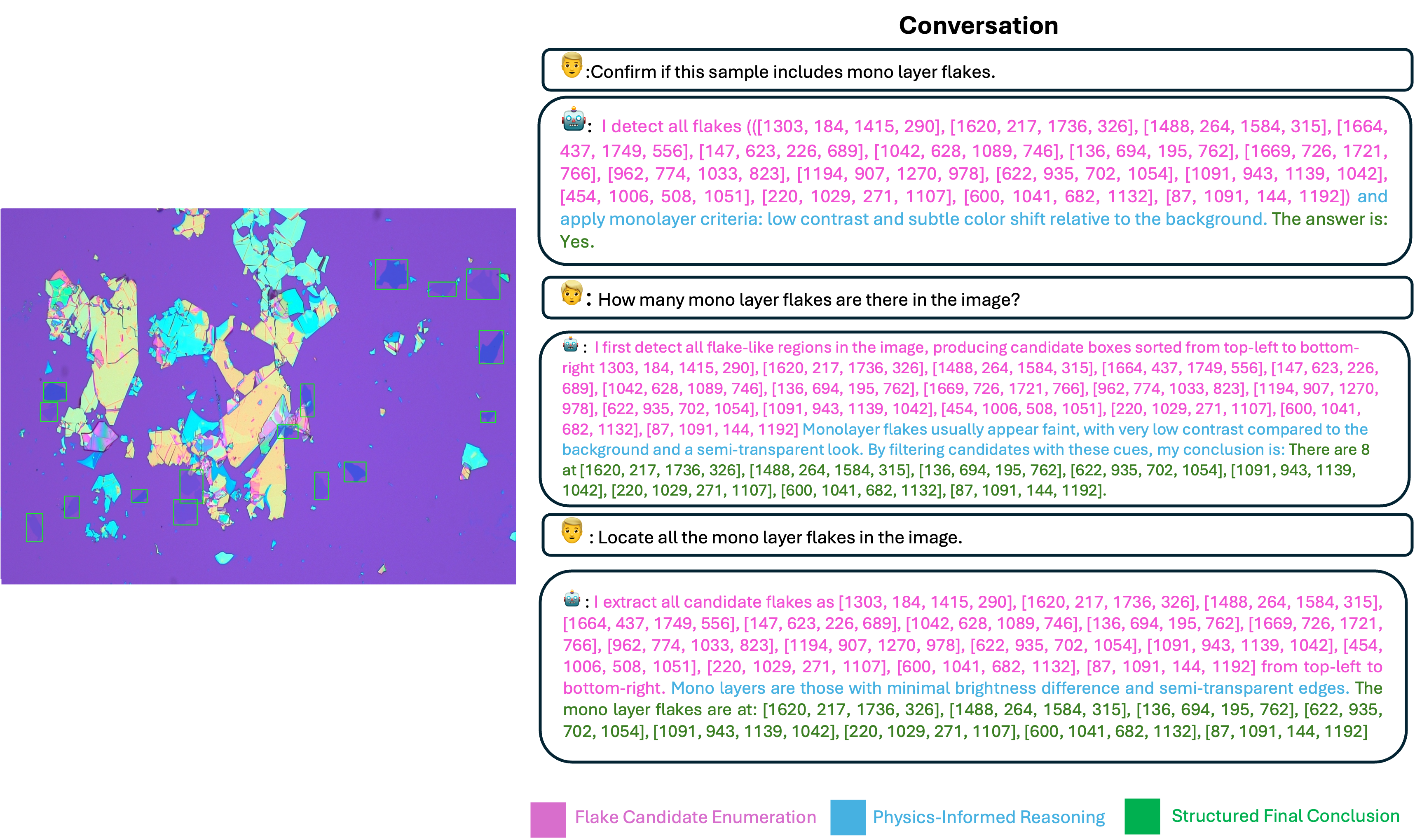}
    \caption{Conversation between a human and model analyzing a synthetic optical microscopy image. The model output is highlighted in three parts: flake candidate enumeration (pink), physics-informed reasoning (blue), and final conclusion (green).}
    \label{fig:synth_conversation6}
\end{figure*}

\begin{figure*}
    \centering
    \includegraphics[width=1\linewidth]{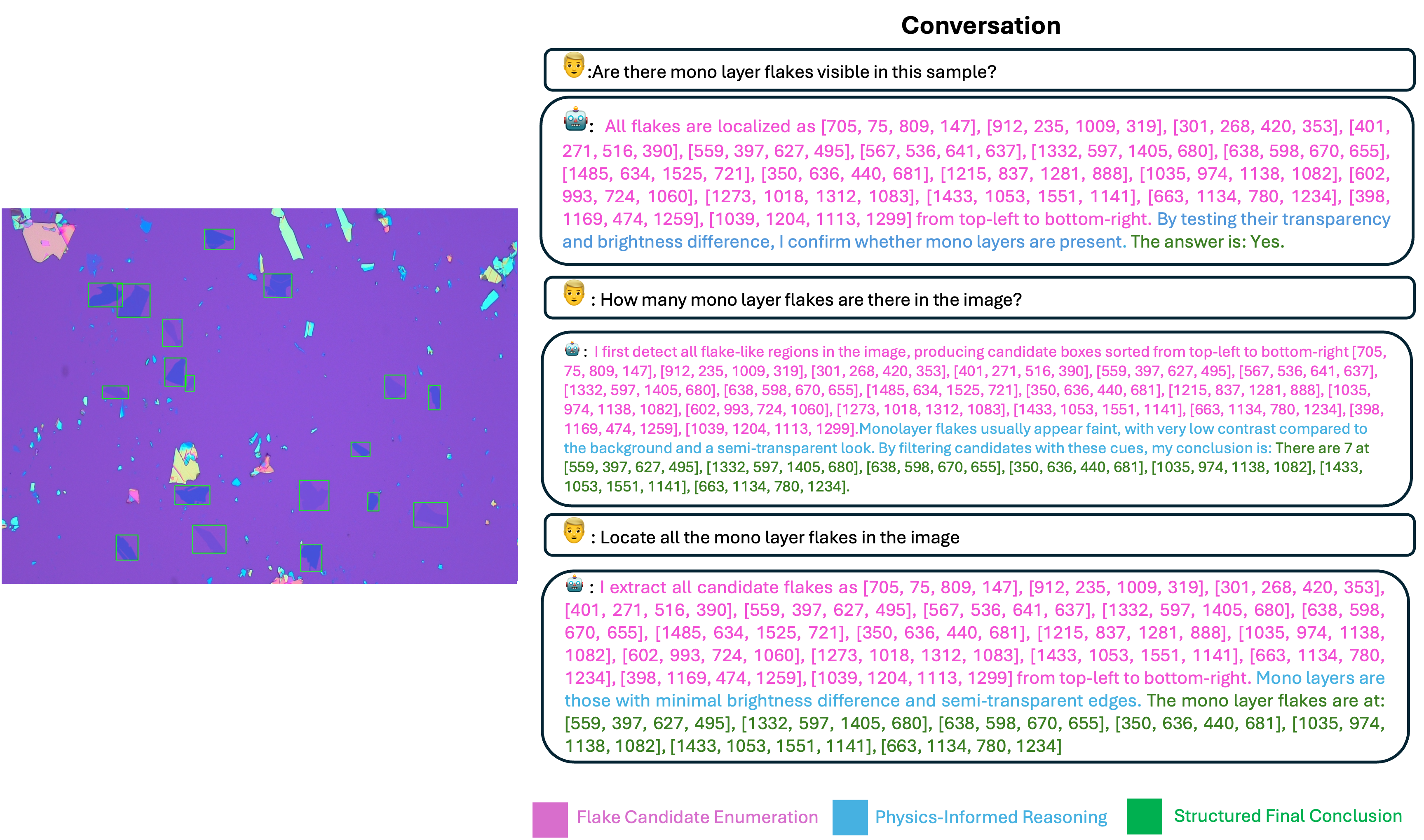}
    \caption{Conversation between a human and model analyzing a synthetic optical microscopy image. The model output is highlighted in three parts: flake candidate enumeration (pink), physics-informed reasoning (blue), and final conclusion (green).}
    \label{fig:synth_conversation7}
\end{figure*}

\begin{figure*}
    \centering
    \includegraphics[width=1\linewidth]{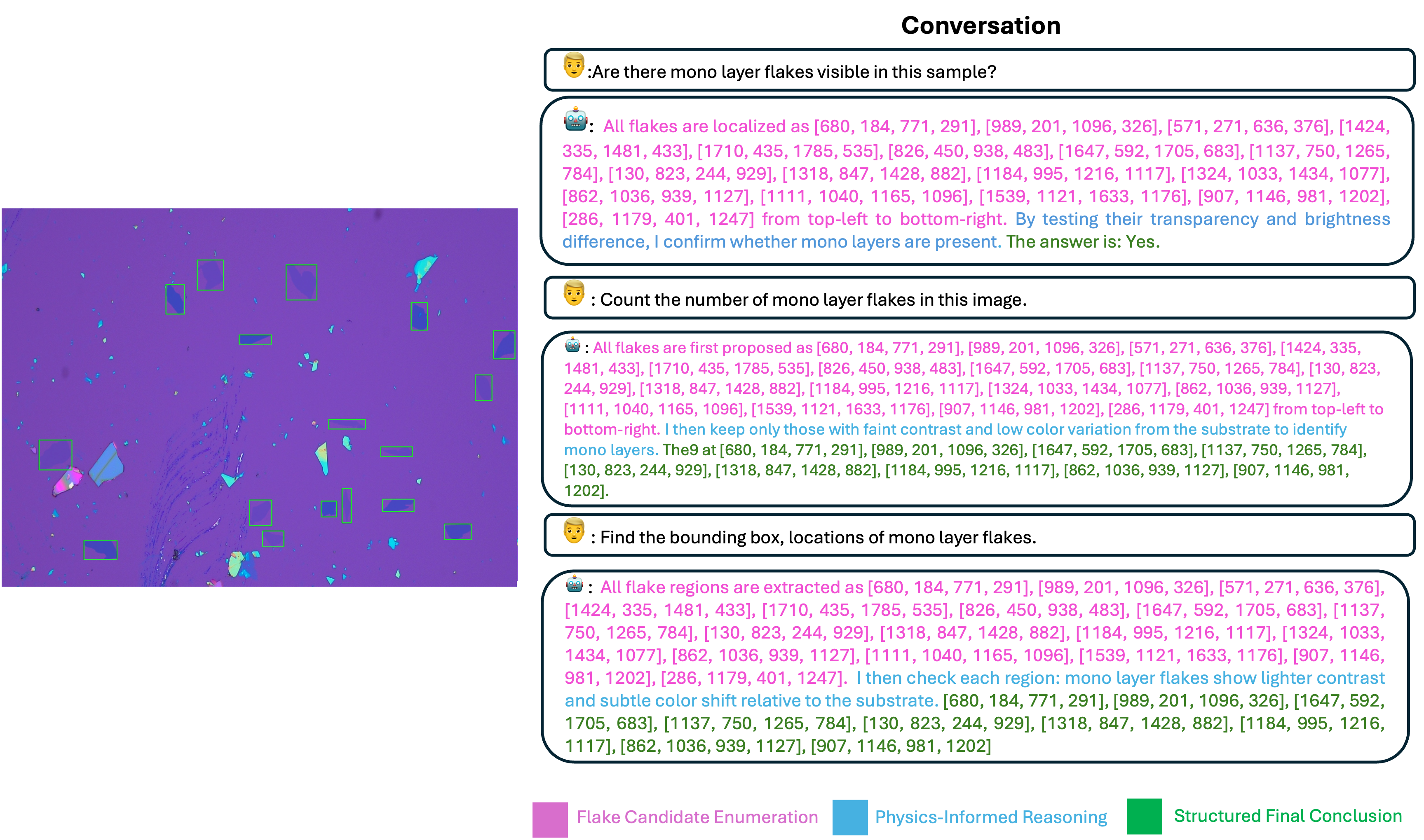}
    \caption{Conversation between a human and model analyzing a synthetic optical microscopy image. The model output is highlighted in three parts: flake candidate enumeration (pink), physics-informed reasoning (blue), and final conclusion (green).}
    \label{fig:synth_conversation8}
\end{figure*}

\begin{figure*}
    \centering
    \includegraphics[width=1\linewidth]{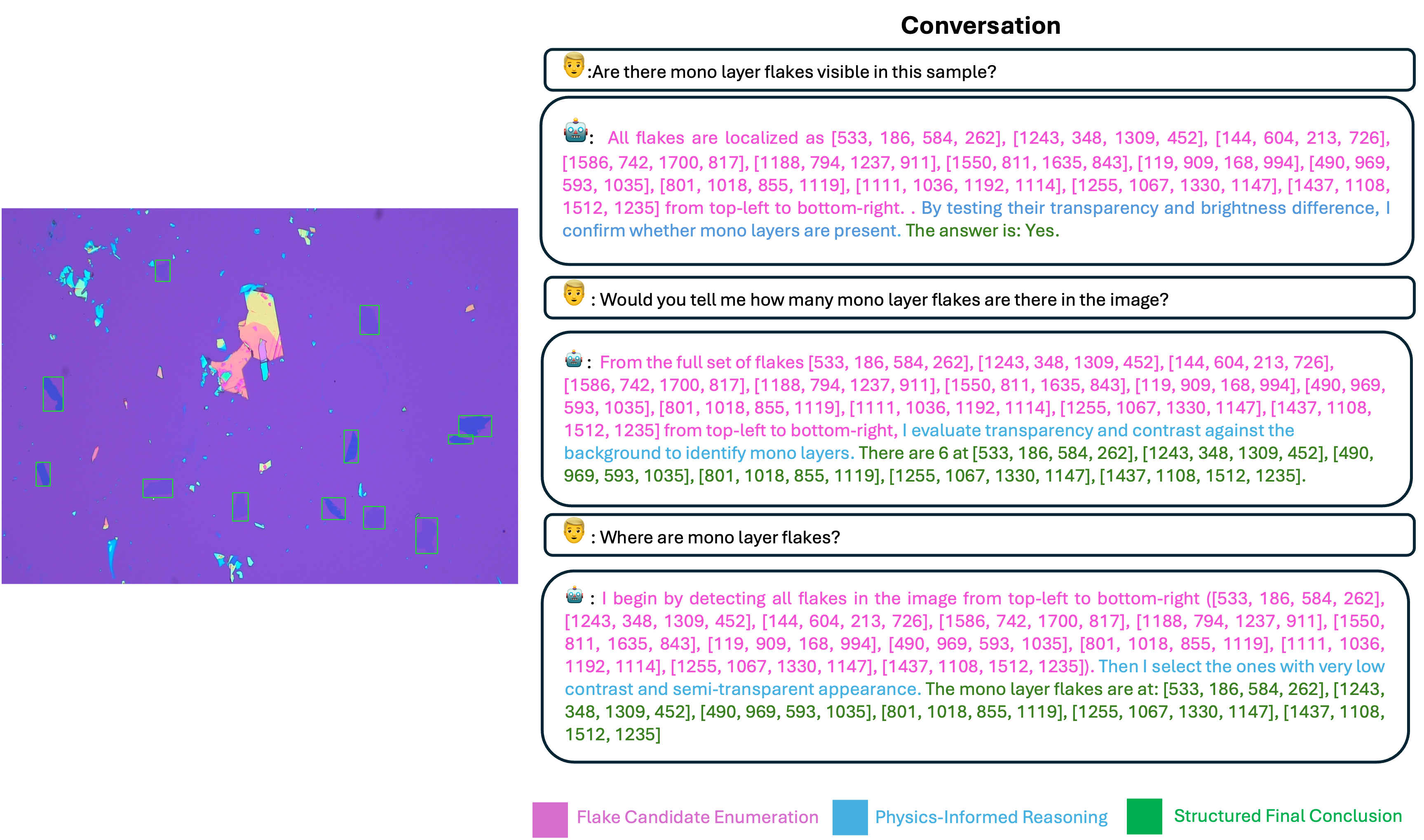}
    \caption{Conversation between a human and model analyzing a synthetic optical microscopy image. The model output is highlighted in three parts: flake candidate enumeration (pink), physics-informed reasoning (blue), and final conclusion (green).}
    \label{fig:synth_conversation9}
\end{figure*}

\begin{figure*}
    \centering
    \includegraphics[width=1\linewidth]{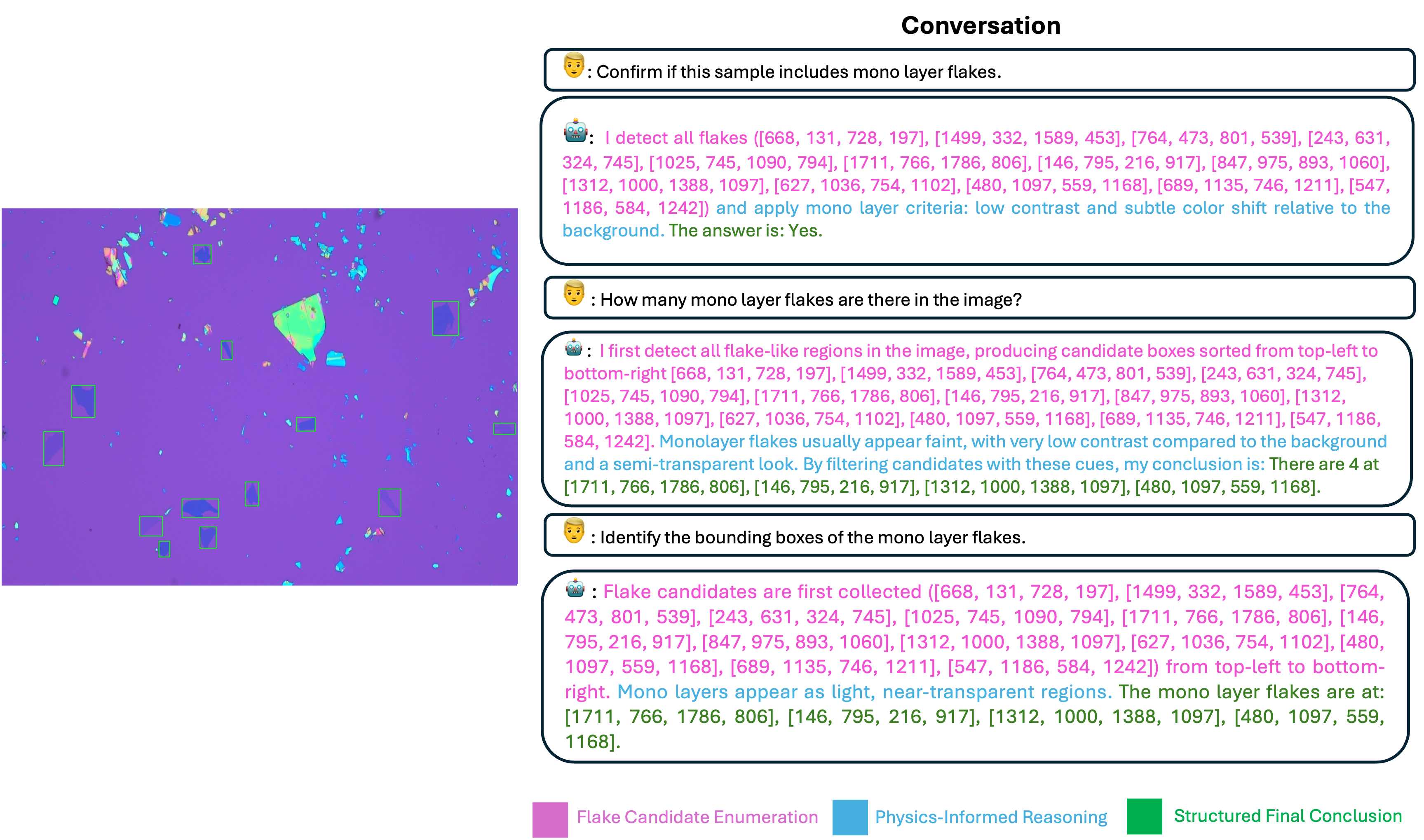}
    \caption{Conversation between a human and model analyzing a synthetic optical microscopy image. The model output is highlighted in three parts: flake candidate enumeration (pink), physics-informed reasoning (blue), and final conclusion (green).}
    \label{fig:synth_conversation10}
\end{figure*}

We present samples from the QMat-Instruct dataset in ~\cref{fig:synth_conversation1} through ~\cref{fig:synth_conversation10}, showing the dialogue between a human and the model concerning material analysis using synthetic optical microscopy images.

\section{QF-Bench Dataset}
\begin{figure*}
    \centering
    \includegraphics[width=1\linewidth]{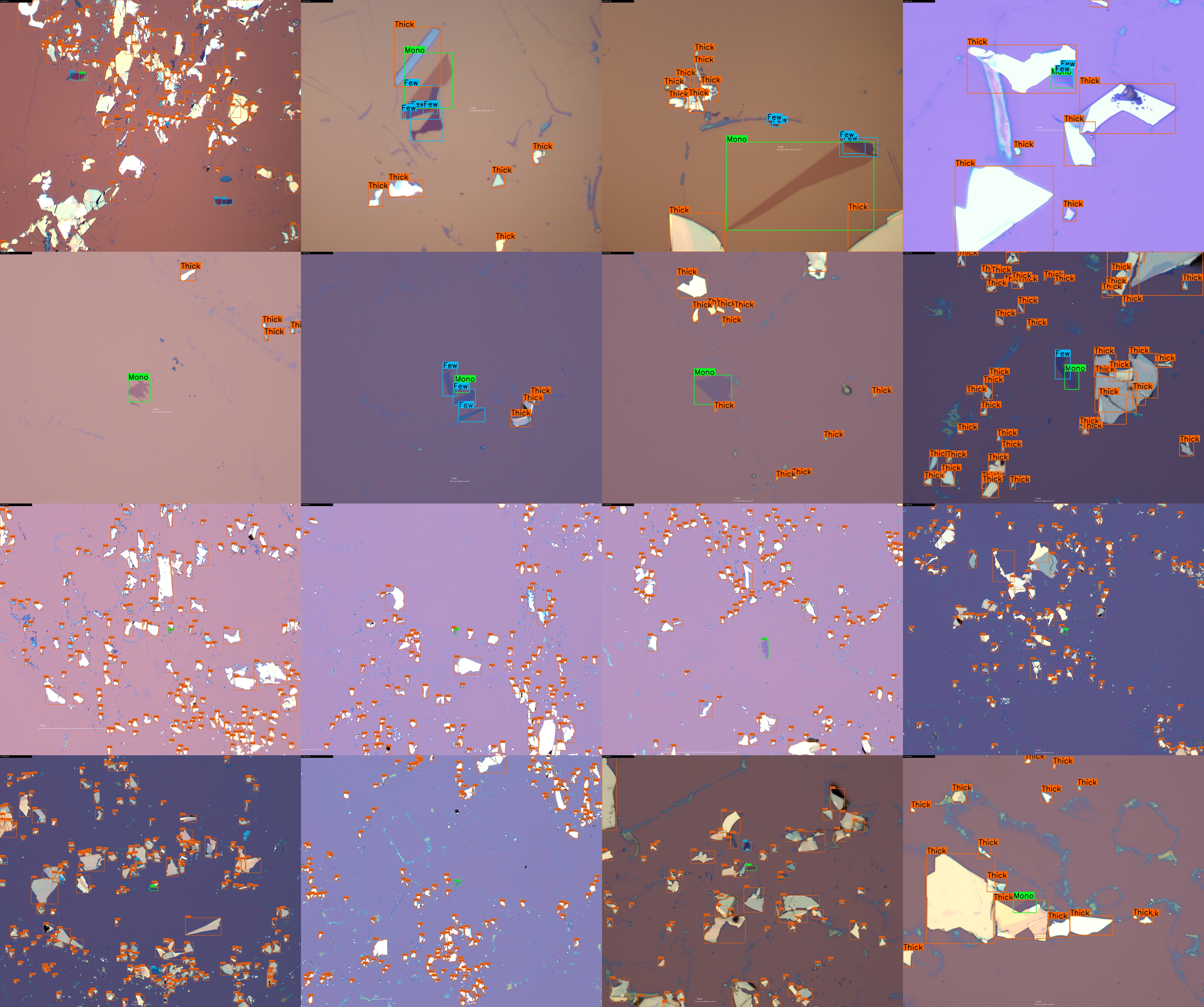}
    \caption{Samples from the QF-Bench dataset. Each microscopy image displays annotations for 2D material flakes. Flake thickness categories are color-coded: green (\texttt{Mono}), blue (\texttt{Few}), and orange (\texttt{Thick}).}
    \label{fig:combined_benchmark}
\end{figure*}

Due to the large size of the data, we currently release 50 samples from the QF-Bench dataset. We show some samples from the benchmark dataset in ~\cref{fig:combined_benchmark}.
The benchmark release is structured in COCO JSON format. The JSON contains top-level keys such as \texttt{images}, \texttt{annotations}, and \texttt{categories}. The dataset defines three distinct categories based on flake thickness: \texttt{Mono}, \texttt{Few}, and \texttt{Thick}, all grouped under the \texttt{flake} supercategory. Each individual flake has a bounding box annotation.

\end{document}